\documentclass{article}

\usepackage{microtype}
\usepackage{graphicx}
\usepackage{subcaption}
\usepackage{booktabs} 

\usepackage{hyperref}

\usepackage[accepted]{icml2023}

\usepackage{amsmath}
\usepackage{amssymb}
\usepackage{mathtools}
\usepackage{amsthm}
\usepackage{bbm}

\usepackage[capitalize,noabbrev]{cleveref}

\theoremstyle{plain}

\theoremstyle{definition}

\theoremstyle{remark}

\usepackage{natbib}
\usepackage{multirow}
\usepackage{graphicx}
\usepackage[textsize=tiny]{todonotes}

\DeclareMathOperator*{\argmax}{argmax}

\usepackage{enumitem}

\newtheorem{innercustomgeneric}{\customgenericname}
\providecommand{\customgenericname}{}
\newcommand{\newcustomtheorem}[2]{%
  \newenvironment{#1}[1]
  {%
   \renewcommand\customgenericname{#2}%
   \renewcommand\theinnercustomgeneric{##1}%
   \innercustomgeneric
  }
  {\endinnercustomgeneric}
}

\newcustomtheorem{customthm}{\bf{Theorem}}
\newcustomtheorem{customlemma}{\bf{Lemma}}

\icmltitlerunning{Dual Focal Loss for Calibration}

\begin{document}

\twocolumn[
\icmltitle{Dual Focal Loss for Calibration}



\icmlsetsymbol{equal}{*}

\begin{icmlauthorlist}
\icmlauthor{Linwei Tao}{equal,yyy}
\icmlauthor{Minjing Dong}{equal,yyy}
\icmlauthor{Chang Xu}{yyy}
\end{icmlauthorlist}

\icmlaffiliation{yyy}{School of Computing Science, University of Sydney}

\icmlcorrespondingauthor{Linwei Tao}{linwei.tao@sydney.edu.au}
\icmlcorrespondingauthor{Chang Xu}{c.xu@sydney.edu.au}

\icmlkeywords{Machine Learning, ICML}

\vskip 0.3in
]



\printAffiliationsAndNotice{\icmlEqualContribution} 

\begin{abstract}

The use of deep neural networks in real-world applications require well-calibrated networks with confidence scores that accurately reflect the actual probability. However, it has been found that these networks often provide over-confident predictions, which leads to poor calibration. Recent efforts have sought to address this issue by focal loss to reduce over-confidence, but this approach can also lead to under-confident predictions. While different variants of focal loss have been explored, it is difficult to find a balance between over-confidence and under-confidence. In our work, we propose a new loss function by focusing on dual logits. Our method not only considers the ground truth logit, but also take into account the highest logit ranked after the ground truth logit. By maximizing the gap between these two logits, our proposed dual focal loss can achieve a better balance between over-confidence and under-confidence. We provide theoretical evidence to support our approach and demonstrate its effectiveness through evaluations on multiple models and datasets, where it achieves state-of-the-art performance. Code is available at https://github.com/Linwei94/DualFocalLoss
\end{abstract}

\section{Introduction}

It is well-established that Deep Neural Networks (DNNs) have achieved exceptional results in computer vision tasks, such as image classification \citep{he2016deep,zagoruyko2016wide}, object detection \citep{he2017mask,tian2019fcos}, and semantic segmentation \citep{long2015fully,cheng2021per}. The focus has primarily been on achieving accurate predictions for state-of-the-art performance. However, the reliability of the predicted confidence scores, also known as uncertainty, is just as important when deploying models in real-world applications. Despite their high accuracy, most DNNs struggle to accurately reflect the actual probabilities of their predictions through their confidence scores. For instance, if a DNN assigns a confidence score of $0.8$ to a set of predictions, it should be correct $80\%$ of the time. However, DNNs tend to overestimate the actual probability of correct predictions, making it difficult for downstream components to rely on them. Therefore, calibrating the uncertainty of DNNs can be crucial for their successful deployment in real-world applications.

In recent years, there has been increasing attention on understanding the causes of poor calibration in deep neural networks (DNNs) and finding ways to improve it. One main focus has been on the role of the loss function, as it has been shown to have a strong influence on calibration. Prior work has proposed various techniques to improve calibration by adding a calibration regularization term to the cross-entropy loss, such as Maximum Mean Calibration Error (MMCE)~\citep{kumar2018trainable}, AvUC~\citep{krishnan2020improving}, Soft Calibration Objective~\citep{karandikar2021soft} and Meta Calibration~\citep{bohdal2021meta}, and others have proposed alternatives like Brier loss~\citep{hui2020evaluation}. It has also been shown that replacing Cross-Entropy (CE) loss with Focal Loss (FL) can improve calibration performance~\citep{mukhoti2020calibrating}. The inverse focal loss has also been proposed to preserve sample hardness information, which benefits post-hoc calibration methods~\citep{wang2021rethinking}. Although variants of focal loss are proposed to improve calibration performance by alleviating the over-confidence problems, there exist trade-offs between over-confidence and under-confidence in loss function design, which existing techniques cannot tackle. For example, focal loss achieves better calibration error through guiding networks to provide predictions with higher entropy, which leads to under-confidence. On the contrary, inverse focal loss aims at preserving the sample hardness while aggravating the over-confidence issue. Neither over-confident nor under-confident predictions are feasible for downstream components, which motivates us to explore this trade-off through the lens of loss function design. 


Current loss functions for calibrating DNNs typically rely on a single logit corresponding to the ground truth label, while the impact of other logits on calibration has yet to be thoroughly studied. This can make it difficult for the loss function to identify and address over-confidence and under-confidence scenarios. This paper introduces a new loss function called Dual Focal Loss (DFL), which takes into account the logit corresponding to the ground truth label and the largest logit ranked after it. By maximising the gap between these dual logits, DFL aims to find a better balance between over-confidence and under-confidence automatically. To demonstrate the effectiveness of DFL, we provide theoretical evidence of its ability to improve calibration through instance-wise conditional risk analysis. By analysing over-confidence and under-confidence regions using the proposed dual logits, we show that DFL can significantly reduce the size of the under-confidence region while preserving the advantages of FL in over-confidence scenarios, resulting in improved overall calibration performance.

Our main contributions in this work can be summarized as follows:
\textbf{(1)} We propose a new formulation of focal loss for calibration that uses dual logits.
\textbf{(2)} We provide theoretical analysis to support the superiority of the proposed Dual Focal Loss (DFL).
\textbf{(3)} We perform extensive evaluations on multiple datasets and models, and our proposed DFL achieves state-of-the-art calibration performance.

\section{Related Work}
Many techniques have been proposed in recent years to address the network miscalibration problem. These methods can be divided into three categories. The first category is post-hoc calibration techniques that adjust model predictions after training by optimizing additional parameters on a held-out validation set. These methods include Platt Scaling~\citep{platt1999probabilistic}, which learns parameters to perform a linear transformation on the original prediction logits, Isotonic Regression~\citep{zadrozny2002transforming}, which learns piece-wise functions to transform the original prediction logits, Histogram Binning~\citep{zadrozny2001obtaining}, which obtains calibrated probability estimates from decision trees and naive Bayesian classifiers, Bayesian Binning into Quantiles (BBQ)~\citep{naeini2015obtaining}, an extension of histogram binning with Bayesian model averaging, Beta calibration~\citep{kull2017beta}, which is proposed for binary classification and generalized to multi-classification with Dirichlet distributions by~\citet{kull2019beyond}. Temperature scaling~\citep{guo2017calibration} is a widely used post-hoc calibration method, which maximizes the temperature parameter in the SoftMax function on held-out negative log-likelihood. In this work, we report the calibration performance on multiple metrics together with the post-temperature scaling results.

The second category of calibration methods is those related to regularization. Regularization techniques are generally found to be effective for calibrating DNNs. Data augmentation methods, such as Mixup~\citep{thulasidasan2019mixup} and AugMix~\citep{hendrycks2019augmix}, train DNNs on mixed samples and have been found to reduce the tendency to make over-confident predictions. Model ensemble techniques, where multiple DNNs are trained individually and their predictions are averaged, successfully improve both the accuracy and predictive uncertainty of a single DNN. This is because ensemble methods can reduce the over-confidence in predictions by aggregating the outputs of multiple models. The effect of ensemble methods on calibration has been extensively studied by~\citet{lakshminarayanan2017simple, zhang2020mix, rahaman2021uncertainty}. Batch Ensemble~\citep{wen2020batchensemble} provides an efficient way to train a calibrated network. Label smoothing~\citep{muller2019does} is a technique that implicitly calibrates DNNs by replacing one-hot encoded labels with softened targets, encouraging networks to produce more uncertain predictions and thereby reducing over-confidence. Weight decay, which regularizes networks by penalizing weights based on their L2 norm, is also shown to be effective for confidence calibration~\citep{guo2017calibration}.

The third category of calibration methods is those that modify the training loss to improve calibration. These methods include adding a differentiable auxiliary surrogate loss for expected calibration error, such as in~\citep{karandikar2021soft, kumar2018trainable, krishnan2020improving, bohdal2021meta}, or replacing the training loss with other loss functions such as mean square error loss~\citep{hui2020evaluation}, inverse focal loss~\citep{wang2021rethinking} and focal loss~\citep{gupta2020calibration}. Among these methods, focal loss~\citep{gupta2020calibration}, which adds a modulating term to the cross-entropy loss to focus learning on hard examples, is a simple and efficient way to train a calibrated model. The focal loss and DFL can be categorize into margin-based loss like Hinge Loss\cite{cortes1995support}, Contrastive Loss\cite{hadsell2006dimensionality}, Triplet Loss\cite{schroff2015facenet} and Margin Ranking Loss~\cite{weston2011wsabie}.

Expected Calibration Error (ECE)~\citep{guo2017calibration} is a widely accepted metric in the literature. However, recent works~\citep{nixon2019measuring,kumar2019verified,roelofs2022mitigating,gupta2020calibration} point out the limitations of ECE, such as its sensitivity to bin size. For a fair comparison, we report the results in terms of four calibration metrics, i.e., ECE, Maximum Calibration Error (MCE)~\citep{guo2017calibration}, Adaptive-ECE~\citep{ding2020revisiting} and classwise-ECE~\citep{kull2019beyond} and also provide reliability diagrams~\citep{niculescu2005predicting} for visual comparison.

\section{Methodology}
\subsection{Problem Formulation}
Considering a classification task with $\mathcal{X}$ as the input space and $\mathcal{Y}$ as the label space, the classifier $q$ maps $x \in \mathcal{X}$ to a probability distribution $\hat{p}$ on $K$ classes and $k = \argmax_i \hat{p}_i$ denotes the index of the predicted label. The ground-truth $y \in \mathcal{Y}$ and predicted labels $\hat{y} \in \mathcal{Y}$ are formulated in one-hot format where $y_{gt} = 1$ and $\hat{y}_k = 1$. Then, the associated confidence score of the predicted label is $\hat{p}_k$.

\noindent\textbf{Classification Calibration}
The network is said to be perfectly calibrated if the predicted confidence $\hat{p}$ presents the true probability that the classification is correct. Formally, the perfectly calibrated network satisfies $\mathbb{P}(\hat{y}=y|\hat{p}=p) = p$ for all $p \in [0,1]$ \citep{guo2017calibration}. Given the confidence score and the probability of correctness, the \textit{Expected Calibration Error} (ECE) is defined as
$\mathbb{E}_{\hat{p}}[|\mathbb{P}(\hat{y}=y|\hat{p})-\hat{p}|]$. In practice, since the calibration error cannot be derived due to the finite samples in the datasets, an approximation of ECE is introduced in \citep{guo2017calibration}. Specifically, all the samples are grouped into $M$ bins $\{B_m\}^M_{m=1}$ with the same interval according to their confidence scores, where $B_m$ contains all the samples with their confidence scores $\hat{p}_k \in [\frac{m}{M},\frac{m+1}{M})$. For each bin $B_m$, the average confidence is computed as $C_m = \frac{1}{|B_m|}\sum_{i \in B_m} \hat{p}_k^i$ and the bin accuracy is computed as $A_m = \frac{1}{|B_m|}\sum_{i\ in B_m} \mathbbm{1}(\hat{y}^i_k = y^i_k)$ where $\mathbbm{1}$ is the indicator function. Then the ECE can be approximated as the expected absolute difference between bin accuracy and average confidence as
\begin{equation} \label{eq:ece}
    \text{ECE} = \sum^M_{m=1}\frac{|B_m|}{N}|A_m-C_m|,
\end{equation}
where $N$ denotes the number of samples. Besides the estimated ECE in Eq. \ref{eq:ece}, there exist ECE variants to measure this error. For example, AdaECE~\citep{nguyen2015posterior} group samples into the bins $B_m$ with equal number of samples where $|B_m| = |B_n|$, and ClasswiseECE~\citep{kull2019beyond} approximates the ECE over $K$ classes.

\noindent\textbf{Temperature Scaling} A popular technique to tackle the classification calibration is temperature scaling, which adjusts the sharpness of output probability distribution via the temperature in the SoftMax function as $\hat{p}_i = \frac{exp(\hat{g}_i/T)}{\sum^K_{k=1} exp(\hat{g}_k/T)}$, where $\hat{g}$ denotes the logits before SoftMax function. The model calibration performance can be improved by tuning $T$ on a hold-out validation set.

\subsection{Dual Focal Loss for Calibration}
Focal loss is originally introduced to tackle the foreground and background imbalance in object detection by assigning larger weights to hard samples and smaller weights to easy samples. Formally, the focal loss is defined as
\begin{equation} \label{eq:FL}
    \mathcal{L}_{FL}(x,y) = -\sum^K_{i=1} y_i(1-q_i(x))^{\gamma}\text{log }q_i(x),\\
\end{equation}
where $\gamma$ is the pre-defined hyperparameter and $q$ denotes the score function. It is easy to see that the focal loss with $\gamma = 0$ is equivalent to the cross-entropy loss $\mathcal{L}_{CE}(x,y) = -\sum^K_{i=1} y_i\text{log }q_i(x)$. It was shown that the models optimized by focal loss perform better calibration than cross-entropy loss in prior work~\citep{mukhoti2020calibrating}. They mainly attributed the improvement to the fact that focal loss has the effect of adding a maximum-entropy regularizer as $\mathcal{L}_{FL} \geq \text{KL}(y||\hat{p}) - \gamma \text{H}(\hat{p})$ where H denotes the entropy of prediction $\hat{p}$. With this regularizer, the predicted distribution is optimized to have higher entropy to tackle over-confidence with cross-entropy loss.
However, there exists a major concern about using focal loss in calibration. The focal loss could suffer from under-confidence since it penalizes the all the output probability logits to a low level~\citep{charoenphakdee2021focal}, and predictions are optimized to form a tight probability distribution without distinction, which leads to the loss of the sample hardness information~\citep{wang2021rethinking}. Although the conservative predictions guided by the focal loss have a smaller ECE, the lower confidence scores may not accurately reflect the actual probability. An inverse version of focal loss is proposed to preserve the hardness. However, it aggravates the over-confidence issue in DNNs~\citep{wang2021rethinking}. Thus, a trade-off exists between over-confidence and under-confidence in the loss function design for calibration. For a better trade-off, we take the focal loss as a strong baseline and propose to modify it to achieve two objectives: 1). Alleviate the under-confidence issue in focal loss and encourage the networks to provide courageous but accurate confidence scores; 2). Preserve the superiority of focal loss in the over-confidence scenario.

We mainly attribute the failure cases of previous work to the fact that only $q_{gt}(x)$ is involved in the computation, where $gt$ denotes the index of the ground truth label. For example, cross-entropy loss forces $q_{gt}(x)$ to be close to $1$ and focal loss forces $q_{gt}(x)$ to be low confidence. However, there are no direct connections between $q_{gt}(x)$ and other logits $q_{i}(x), i\neq gt$ to indicate the scenario of over/under-confidence in the loss function. For example, if the ground truth class probability $q_{gt}(x)$ is fixed at 0.5, a loss function will provide the same loss regardless of whether the remaining confidence is evenly distributed among the other logits, or if one logit is equal to 0.49. Thus, we argue that $q_{gt}(x)$ is not enough for the loss function in calibration. In this work, we propose reformulating Eq. \ref{eq:FL} by considering dual logits to tackle the abovementioned issues. Formally, the dual focal loss for calibration is defined as
\begin{equation} \label{eq:DFL}
\begin{aligned}
    & \mathcal{L}_{DFL} = -\sum^K_{i=1} y_i(1-q_i(x)+q_j(x))^{\gamma}\text{log }q_i(x),\\
    & \text{where } q_j(x) = \max_i \{q_i(x)|q_i(x) < q_{gt}(x)\}.\\
\end{aligned}
\end{equation}
The major difference between FL and DFL lies in the involvement of $q_j(x)$, which denotes the maximum value in the descending order $q_{1:K}(x)$ after $q_{gt}(x)$. Intuitively, Eq. \ref{eq:DFL} inherits the form in focal loss to prevent over-confidence while aiming at maximizing the gap between dual logits $q_{gt}(x)$ and $q_{j}(x)$ to encourage the networks to enlarge their confidence scores if possible. With the proposed dual logits in Eq. \ref{eq:DFL}, a better trade-off between over/under-confidence can be achieved, and the calibration performance can be further improved, which is empirically verified in Sec. \ref{sec:exp} and theoretically analyzed in Sec. \ref{sec:theoretical}.

\section{Theoretical Evidence} \label{sec:theoretical}
In this section, we provide theoretical evidence that our proposed DFL can be more effective for confidence calibration than FL. We first examine the instance-wise conditional risk to reveal the relationship between the risk minimizer and the actual class-posterior probability. We then demonstrate that DFL is a classification-calibrated loss. Finally, we highlight the superiority of DFL by showing that it significantly reduces the under-confident region compared to FL.

\subsection{Instance-wise Conditional Risk} 
Following~\citep{bartlett2006convexity,tewari2007consistency}, we first define the instance-wise conditional risk $\mathcal{R}$ for both FL and DFL in Eqs. \ref{eq:FL} and \ref{eq:DFL} as
\begin{equation}
\begin{aligned}
    & \mathcal{R}_{FL} = - \sum^K_{i=1} \eta_i(x) (1-q_i(x))^{\gamma}\text{log }q_i(x),\\
    & \mathcal{R}_{DFL} = - \sum^K_{i=1} \eta_i(x) (1-q_i(x)+q_j(x))^{\gamma}\text{log }q_i(x),\\
\end{aligned}
\end{equation}
where $x$ denotes the data point, and $\eta$ denotes the actual class-posterior probability, which corresponds to the well-calibrated confidence score of the classifier. Thus, $\mathcal{R}$ denotes the expected penalty for a data point $x$ with $q$ as the score function. Through exploring the instance-wise conditional risk $\mathcal{R}$, the relationship between $q$ and $\eta$ can be derived to indicate the influence of loss function on classification calibration. The instance-wise conditional risk of focal loss has been well-studied in~\citep{charoenphakdee2021focal}. In this work, we generalize the results to our proposed DFL and highlight its superiority over FL via theoretical evidence. We omit the usage of $x$ for brevity. Formally, we consider a general form of dual focal loss where we use $f(q_i)$ to represent $q_j(x)$ in Eq. \ref{eq:DFL} and the optimization of $\mathcal{R}_{DFL}$ can be formulated as
\begin{equation} \label{eq:risk_obj}
\begin{aligned}
    & \min_{q} - \sum^K_{i=1} \eta_i(1-q_i+f(q_i))^{\gamma}\text{log}\;q_i, \\
    & \text{subject to } \sum^K_{i=1} q_i=1.
\end{aligned}
\end{equation}
To tackle the constrained optimization in Eq. \ref{eq:risk_obj}, we consider the following Lagrangian equation:
\begin{equation}
\scalebox{0.96}{$ \displaystyle
\begin{aligned}
    & \mathcal{L}(q,\lambda) = - \sum^K_{i=1} \eta_i(1 - q_i + f(q_i))^{\gamma}\text{log}\;q_i + \lambda \Bigl(\sum^K_{i=1} q_i - 1 \Bigl),
\end{aligned}
$}
\end{equation}
where $\lambda$ is the Lagrangian multiplier for the constraint. Through taking the derivatives with respect to $q_i$, $\lambda$ can be solved at the optimal $q^*$ as
\begin{equation} \label{eq:Lagrangian}
\begin{aligned}
    & \frac{\partial}{\partial_{q_{i}}}\mathcal{L}(q,\lambda) \Bigl|_{q=q^*} = 0, \\
    & (1 - \frac{\partial f}{\partial q^*_i}) \eta_i \gamma (1-q^*_i+f(q^*_i))^{\gamma-1} \text{ log }q^*_i \\
    & - \eta_i\frac{(1-q^*_i+f(q^*_i))^{\gamma}}{q^*_i} + \lambda= 0, \\
    & \lambda = \frac{\eta_i}{q^*_i} \Big( (\frac{\partial f}{\partial q^*_i} - 1) \gamma (1-q^*_i+f(q^*_i))^{\gamma-1} q^*_i \text{ log }q^*_i\\
    & \;\;\;\;\; + (1-q^*_i+f(q^*_i))^{\gamma} \Big).\\
\end{aligned}
\end{equation}
With Eq. \ref{eq:Lagrangian}, $\eta_i$ can be written as a function of $\lambda$ and $q^*_i$. For simplicity, we let $\phi(v_i) = (1-v_i+f(v_i))^{\gamma} + (\frac{\partial f}{\partial v_i} - 1)\gamma(1-v_i+f(v_i))^{\gamma-1}v_i \text{ log }v_i$. And $\eta_i$ can be solved as $\eta_i = \frac{\lambda q^*_i}{\phi(q^*_i)}$. Since $\eta_i$ follows the probability distribution, $\lambda$ can be rewritten as a function of $q^*_i$ as
\begin{equation} \label{eq:lambda_1}
\begin{aligned}
    & \sum^K_{i} \eta_i = 1 = \lambda \sum^K_i \frac{q^*_i}{\phi(q^*_i)},\\
    & \lambda = \frac{1}{\sum^K_i \frac{q^*_i}{\phi(q^*_i)}}.\\
\end{aligned}
\end{equation}
Similarly, with derived $\lambda$ in Eq. \ref{eq:lambda_1}, $\eta_i$ can be rewritten as a function of $q^*_i$ as
\begin{equation} \label{eq:eta_2}
    \eta_i = \frac{\frac{q^*_i}{\phi(q^*_i)}}{\sum^K_k \frac{q^*_k}{\phi(q^*_k)}}.
\end{equation}

\subsection{Over-confidence and Under-confidence}
We discuss the over-confident and under-confident scenarios of the risk minimizer $q^*$ for DFL. Formally, $\eta$-over/under-confident ($\eta$OC/$\eta$UC) at data point $x$ is defined as
\begin{equation} \label{eq:oc_uc}
\begin{aligned}
    & \max_{i} q_i^{*}(x) -  \max_{i} \eta_i(x) > 0, \text{$q_i^{*}$ is $\eta$OC,}\\
    & \max_{i} q_i^{*}(x) -  \max_{i} \eta_i(x) < 0, \text{$q_i^{*}$ is $\eta$UC.}\\
\end{aligned}
\end{equation}
However, Eq. \ref{eq:oc_uc} cannot be directly simplified via Eq. \ref{eq:eta_2} since it is unknown whether the max-index of $q^{*}$ and $\eta$ are identical. Thus, we first show that our proposed DFL is classification-calibrated to remove $\max$ in Eq. \ref{eq:oc_uc}.

\begin{customthm}{1}\label{thm:DFL}
For any $\gamma > 0$, $\mathcal{L}_{DFL}$ is classification-calibrated and has the strictly order-preserving property.
\end{customthm}
The detailed proof is provided in the supplementary material. Theorem \ref{thm:DFL} indicates that the risk minimizer preserves the order of true class-posterior probability as $q_a^{*}(x) < q_b^{*}(x) \Rightarrow \eta_a(x) < \eta_b(x)$. In other words, the max-index of $q^{*}$ and $\eta$ are identical. Together with Eq. \ref{eq:eta_2}, $\eta$UC of $q^{*}$ in Eq. \ref{eq:oc_uc} can be reformulated as
\begin{equation} \label{eq:oc_ieq}
\begin{aligned}
    & q_m^{*}(x) - \eta_m(x) < 0\\
    & \sum^K_i \frac{q^*_i(x)}{\phi(q^*_i(x))} < \frac{1}{\phi(q^*_m(x))},\\
\end{aligned}
\end{equation}
where $m = \argmax_{i} q^*_i(x)$. Ineq. \ref{eq:oc_ieq} holds if $\phi(q^*_m(x)) \leq \phi(q^*_i(x))$ for all $i \in [1, K]$ and the scenario of $\eta$OC can be explored through flipping the sign of above inequalities. Thus, the monotonicity of function $\phi$ plays an important role in $\eta$OC/$\eta$UC.
We then explore the properties of $\phi$ with DFL. In our proposed DFL, the function $f(v_i)$ maps $v_i$ to the second maximum value $v_j$ in the descending order $v_{1:K}$ after $v_{gt}$, which is defined as
\begin{equation}
\begin{aligned}
    & f(v_i) = v_j,\\
    & \text{where } v_j = \max_i \{v_i|v_i < v_{gt}\}.\\
\end{aligned}
\end{equation}
For simplicity, we denote $v_j=C$. To explore the influence of introduced $f(v_i)$ in $\eta$OC/$\eta$UC, we explore the properties of function $\phi$ in the following Lemma and the detailed proof is provided in the supplementary material.

\begin{figure}[!t]
	\centering
	\includegraphics[width=1\linewidth]{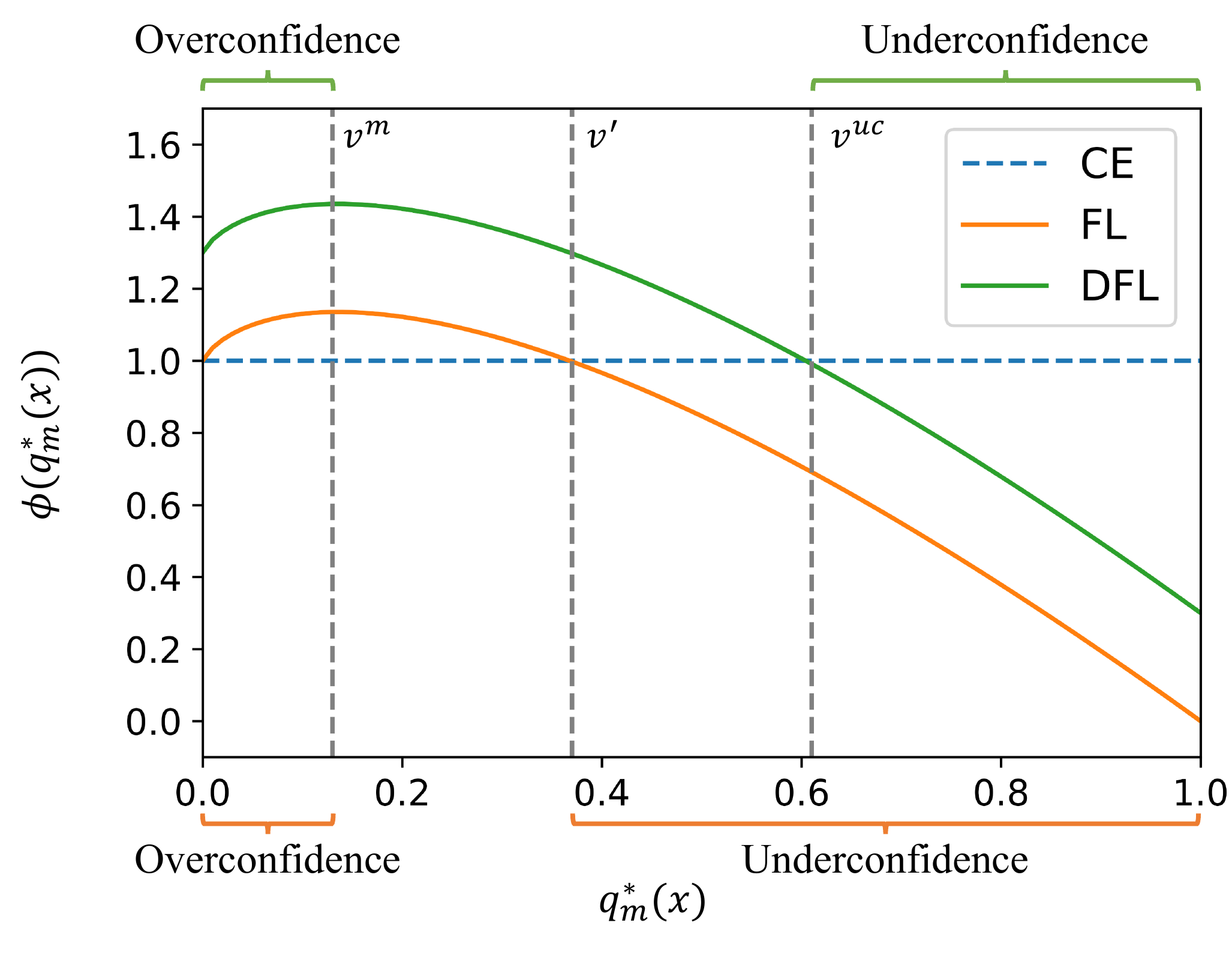}
	\caption{
        \textbf{An illustration of function $\phi$ in FL and DFL.} The over-confident regions of both FL and DFL start from $0.0$ to the maximum point $v^m$. The under-confident region of FL starts from $v'$ to $1.0$ while DFL starts from $v^{uc}$ to $1.0$.
	}
	\label{fig:methodology}
\end{figure}

\begin{customlemma}{1}\label{lemma:phi}
For the scenario where $i \neq j$, $\phi(0) = (1+C)^\gamma \text{ and } \phi(1) = C^{\gamma}$. There exist a unique $v^m \in (0,1)$ such that $\frac{\partial \phi}{\partial v_i} = 0$ at $v^m$,  $\frac{\partial \phi}{\partial v_i} > 0$ for $v \in [0,v^m)$ and $\frac{\partial \phi}{\partial v_i} < 0$ for $v \in (v^m,1]$. For the scenario where $i = j$, $\frac{\partial \phi}{\partial v_i} = 0$ and $\phi(v_j)=1$.
\end{customlemma}
With the explored properties of function $\phi$, it is easy to see that $\phi$ is an increasing function in $[0,v^m)$ and decreasing function in $(v^m,1]$. Then we have
\begin{equation} \label{eq:DFL_min}
\begin{aligned}
    \min_{v_i \in (0, v^{m})} \phi (v_i) =
    \begin{cases}
        \phi(0) = (1+C)^\gamma, & i \neq j\\
        1, & i = j\\
    \end{cases}
\end{aligned}
\end{equation}
According to Eq. \ref{eq:DFL_min}, we can find a unique $v'$ in $(v^{m}, 1]$ where $\phi(v') = (1+C)^{\gamma}$ such that $\min_{v_i \in (0, v']} \phi = \phi(v')$ for $i \neq j$. However, due to the involvement of $f(v_j)$, the $\eta$UC scenario where $\phi(q^*_m(x)) \leq \phi(q^*_j(x))$ does not hold for $[v',1)$ since $\phi(v') = (1+C)^{\gamma} \geq 1 = \phi(v_j)$. Thus, with DFL, the $\eta$UC scenario holds for $[v^{uc},1)$ where $\phi(v^{uc}) = 1$. Similarly, $\eta$OC scenario holds for $(0, v^m]$. Compared with FL, with the assumption of the same $\gamma$ in both DFL and FL, the values of $v^m$ and $v'$ in FL are close to those in DFL, respectively, in practice. For illustration, we visualize function $\phi$ for both FL and DFL with $\gamma=1$ and $q^*_j=0.3$. As shown in Figure \ref{fig:methodology}, x axis denotes $q^*_m(x)$ and y axis denotes $\phi(q^*_m(x))$ in Eq. \ref{eq:oc_ieq}. Thus, both FL and DFL show similar $\eta$OC in range $(0, v^m]$, however, the region sizes of $\eta$UC in FL and DFL are different. Specifically, the $\eta$UC scenario holds for $[v',1)$ in FL and for $[v^{uc},1)$ in DFL so that the $\eta$UC region size is reduced by $\phi^{-1}(1) - \phi^{-1}((1+C)^\gamma)$ in DFL. Thus, our proposed DFL can alleviate the under-confident problem of traditional FL in classification calibration. We empirically verify the effectiveness of this better trade-off in Sec. \ref{sec:trade-off}.

\begin{table*}[!t]
	\centering
	\scriptsize
	\resizebox{\linewidth}{!}{%
		\begin{tabular}{cccccccccccccccccc}
			\toprule
			\textbf{Dataset} & \textbf{Model} & \multicolumn{2}{c}{\textbf{Weight Decay}} &
			\multicolumn{2}{c}{\textbf{Brier Loss}} & \multicolumn{2}{c}{\textbf{MMCE}} &
			\multicolumn{2}{c}{\textbf{Label Smoothing}} & \multicolumn{2}{c}{\textbf{Inverse Focal Loss}} &
			\multicolumn{2}{c}{\textbf{Focal Loss}} &
            \multicolumn{2}{c}{\textbf{Dual Focal}} &\\
			\textbf{} & \textbf{} &
			\multicolumn{2}{c}{\citep{guo2017calibration}} &
			\multicolumn{2}{c}{\citep{brier1950verification}} & \multicolumn{2}{c}{\citep{kumar2018trainable}} &
			\multicolumn{2}{c}{\citep{szegedy2016rethinking}} & \multicolumn{2}{c}{\citep{wang2021rethinking}} &
			\multicolumn{2}{c}{\citep{mukhoti2020calibrating}} &
            \multicolumn{2}{c}{Ours} &\\
			&& Pre T & Post T & Pre T & Post T & Pre T & Post T & Pre T & Post T & Pre T & 
			Post T & Pre T & Post T & Pre T & Post T\\
			\midrule
			
			\multirow{4}{*}{CIFAR-100} & ResNet-50&17.52&3.42(2.1)&6.52&3.64(1.1)&15.32&2.38(1.8)&7.81&4.01(1.1)&17.88&2.98(2.3)&4.5&2.0(1.1)&\textbf{1.08}&\textbf{1.08(1.0)}\\
			& ResNet-110&19.05&4.43(2.3)&7.88&4.65(1.2)&19.14&3.86(2.3)&11.02&5.89(1.1)&19.47&4.52(2.6)&8.56&4.12(1.2)&\textbf{2.90}&\textbf{2.90(1.0)}\\
			& Wide-ResNet-26-10&15.33&2.88(2.2)&4.31&2.7(1.1)&13.17&4.37(1.9)&4.84&4.84(1.0)&16.9&2.28(2.5)&3.03&\textbf{1.64(1.1)}&\textbf{1.79}&1.79(1.0)\\
			& DenseNet-121&20.98&4.27(2.3)&5.17&2.29(1.1)&19.13&3.06(2.1)&12.89&7.52(1.2)&19.42&2.82(2.3)&3.73&\textbf{1.31(1.1)}&\textbf{1.81}&1.81(1.0)\\
			\midrule
			\multirow{4}{*}{CIFAR-10} & ResNet-50&4.35&1.35(2.5)&1.82&1.08(1.1)&4.56&1.19(2.6)&2.96&1.67(0.9)&4.41&1.32(2.8)&1.55&0.95(1.1)&\textbf{0.46}&\textbf{0.46(1.0)}\\
			& ResNet-110&4.41&1.09(2.8)&2.56&1.25(1.2)&5.08&1.42(2.8)&2.09&2.09(1.0)&4.34&\textbf{0.89(2.9)}&1.87&1.07(1.1)&\textbf{0.98}&0.98(1.0)\\
			& Wide-ResNet-26-10&3.23&0.92(2.2)&1.25&1.25(1.0)&3.29&0.86(2.2)&4.26&1.84(0.8)&3.68&0.99(2.7)&1.56&0.84(0.9)&\textbf{0.81}&\textbf{0.81(1.0)}\\
			& DenseNet-121&4.52&1.31(2.4)&1.53&1.53(1.0)&5.1&1.61(2.5)&1.88&1.82(0.9)&4.61&1.07(2.8)&1.22&1.22(1.0)&\textbf{0.57}&\textbf{0.57(1.0)}\\
			\midrule
			Tiny-ImageNet & ResNet-50&15.32&5.48(1.4)&4.44&4.13(0.9)&13.01&5.55(1.3)&15.23&6.51(0.7)&11.51&6.71(1.3)&1.76&1.76(1.0)&\textbf{1.50}&\textbf{1.50(1.0)}\\
           \midrule							
            NLP 20 Newsgroups & Global Pooling CNN&17.92&2.39(2.3)&15.48&6.78(2.1)&13.58&3.22(1.9)&4.79&2.54(1.1)&16.72&2.51(2.1)&6.92&2.19(1.1)&\textbf{1.79}&\textbf{1.79(1.0)}\\
			\bottomrule
		\end{tabular}%
	}
	\caption{\textbf{ECE before and after temperature scaling.}\quad  ECE is measured as a percentage, with lower values indicating better calibration. In the experiments, ECE is evaluated for different methods, both before (pre) and after (post) temperature scaling. The results are calculated with number of bins set as 15. The optimal temperature value, determined on the validation set, is included in brackets.}
	\label{table:ece_tab1}
\end{table*}

\begin{figure*}
    \centering
	\includegraphics[width=\linewidth]{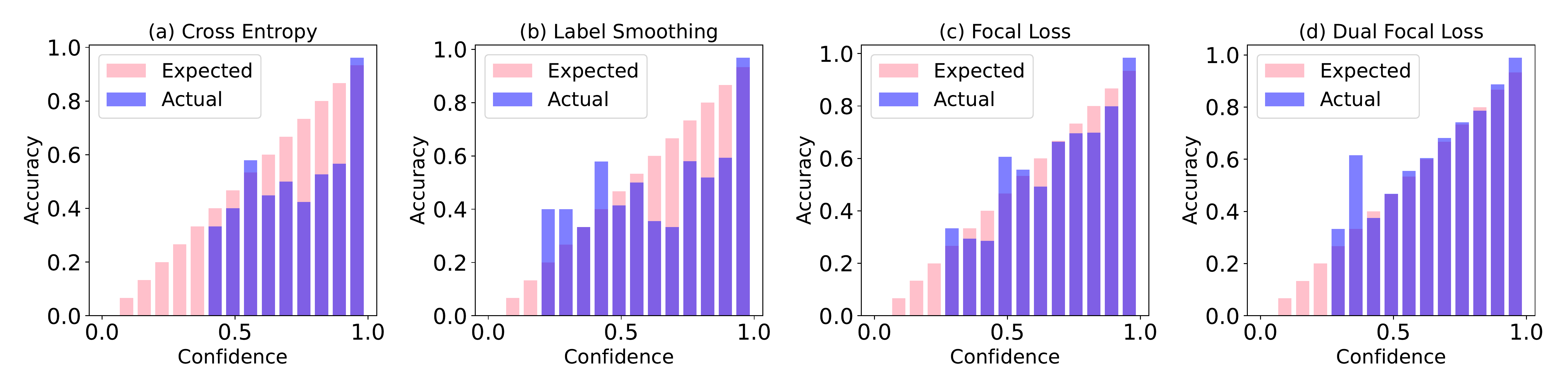}

    \caption{\textbf{Reliability Diagram of Different Methods before Temperature Scaling.}} 
    \label{fig:Reliability Diagram}
\end{figure*}

\section{Experiments} \label{sec:exp}
\noindent
We evaluate our methods on multiple DNNs, including ResNet-50, ResNet-110~\citep{he2016deep}, Wide-ResNet-26-10~\citep{zagoruyko2016wide} and DenseNet-121~\citep{huang2017densely}. Our experiments are conducted on CIFAR-10/100~\citep{krizhevsky2009learning} and Tiny-ImageNet~\citep{deng2009imagenet} for calibration performance. The SVHN dataset, a dataset of street view house numbers and the CIFAR-10-C dataset, a corrupted version of the CIFAR-10 dataset are used as Out-of-Distribution (OoD) datasets for evaluating the robustness of models. The details about datasets can be found in the appendix.

\noindent
\textbf{Baselines} \quad
We compare DFL with multiple approaches, including training with weight decay at $5\times 10^{-4}$ (we find that weight decay at $5\times 10^{-4}$ performs the best), Brier Loss~\citep{brier1950verification}, MMCE loss~\citep{kumar2018trainable}, label smoothing~\citep{szegedy2016rethinking} with a smoothing factor $\alpha_{LS}=0.05$, inverse focal loss~\citep{wang2021rethinking} with $\gamma$ fine-tuned on the split validation set, and focal loss~\citep{mukhoti2020calibrating}.

\noindent
\textbf{Training Setup} \quad
Our training setup follows the prior work~\citep{mukhoti2020calibrating}. We implement our algorithm based on the public code provided by~\citet{mukhoti2020calibrating} and use pre-trained weights provided by ~\citet{mukhoti2020calibrating} for some of the results. We train CIFAR-10/100 for 350 epochs, using 5000 images from the training set for validation. The learning rate is set to $0.1$ for the first 150 epochs, $0.01$ for the following 100 epochs, and $0.001$ for the remaining epochs. For Tiny-ImageNet, we train for 100 epochs, with the learning rate set to $0.1$ for the first 40 epochs, $0.01$ for the following 20 epochs, and $0.001$ for the remaining epochs. We use SGD with a weight decay of $5\times 10^{-4}$ and a momentum of 0.9 for all experiments. The training and testing batch sizes for all datasets are set to 128. We conduct all experiments on a single Tesla V-100 GPU with all random seeds set to 1. For results of temperature scaling, the temperature parameter $T$ is optimized through grid search, with $T\in[0,0.1,0.2,\dots,10]$ on the validation set, based on the best post-temperature-scaling ECE. The exact temperature parameter is used for other metrics, such as adaptive-ECE. Additional details on more experiment results can be found in the appendix.

\begin{figure*}[t]
	\centering
	\includegraphics[width=1\linewidth]{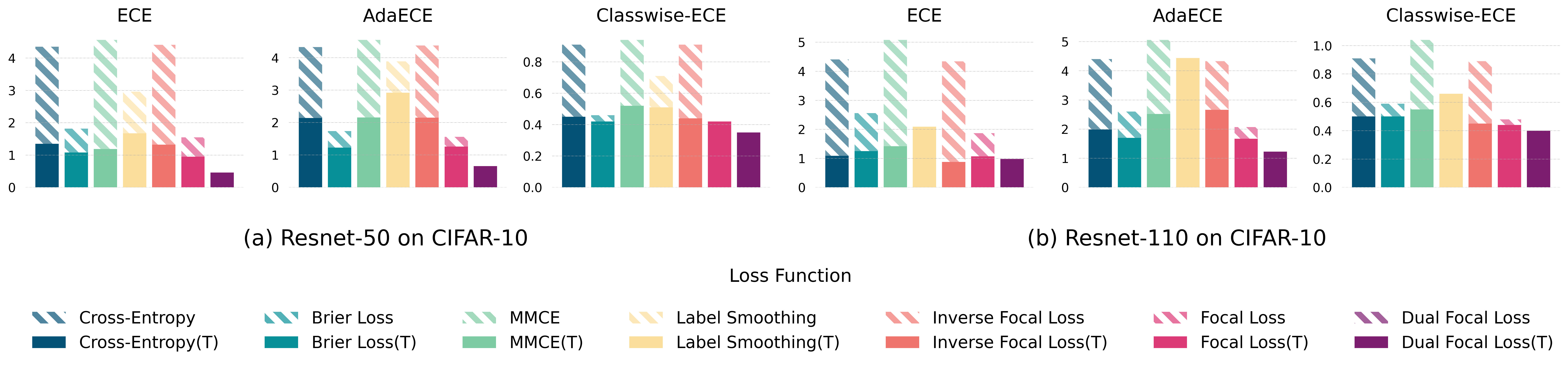}
	\caption{Bar plots with confidence intervals for ECE, AdaECE and Classwise-ECE, computed for ResNet-50 (first 3 figures) and ResNet-110 (last 3 figures) on CIFAR-10.}
	\label{fig:other_metrics}
\end{figure*}

\begin{table*}[!t]
\centering
\scriptsize
\resizebox{\linewidth}{!}{%
\begin{tabular}{ccccccccc}
\toprule
\textbf{Dataset} & \textbf{Model} & \textbf{Weight Decay} &
\textbf{Brier Loss} & \textbf{MMCE} &
\textbf{Label Smoothing} & \textbf{Inverse Focal Loss} &
\textbf{Focal Loss} &
\textbf{Dual Focal} \\
\textbf{} & \textbf{} &
\citep{guo2017calibration} &
\citep{brier1950verification} & \citep{kumar2018trainable} &
\citep{szegedy2016rethinking} & \citep{wang2021rethinking} &
\citep{mukhoti2020calibrating} &
Ours \\

\midrule
\multirow{4}{*}{CIFAR-100} & ResNet-50&23.3&23.39&23.2&23.43&22.23&23.22&22.67\\
& ResNet-110&22.73&25.1&23.07&23.43&22.43&22.51&22.59\\
& Wide-ResNet-26-10&20.7&20.59&20.73&21.19&20.85&20.11&19.91\\
& DenseNet-121&24.52&23.75&24.0&24.05&24.55&22.67&22.4\\
\midrule
\multirow{4}{*}{CIFAR-10} & ResNet-50&4.95&5.0&4.99&5.29&4.80&4.98&5.17\\
& ResNet-110&4.89&5.48&5.4&5.52&4.66&5.42&5.02\\
& Wide-ResNet-26-10&3.86&4.08&3.91&4.2&4.1&4.01&3.96\\
& DenseNet-121&5.0&5.11&5.41&5.09&4.82&5.46&5.43\\
\midrule
Tiny-ImageNet & ResNet-50&49.81&53.2&51.31&47.12&55.19&49.06&48.63\\
\midrule
20 Newsgroups & Global Pooling CNN&26.68	&27.23  &27.06	&26.03	&29.26	&27.98	&28.73\\
\bottomrule
\end{tabular}}
\caption{\textbf{Classification error $(\%)$ on test set for different methods.}}
\label{table:error_tab1}
\end{table*}

\subsection{Calibration Performance}
We report the ECE before and after temperature scaling, along with the optimal temperatures, in Table~\ref{table:ece_tab1}. Our method achieves state-of-the-art ECE in most cases, especially when considering the pre-temperature-scaling results. Our results even substantially exceeded the most post-temperature-scaling results of previous works. The fact that all optimal temperatures for DFL are searched as 1 suggests that DFL trains an innately calibrated model that can achieve good calibration performance without the need for temperature scaling. This is an important factor in developing accurate and reliable models, which are efficient and require less post-processing. The results on CIFAR-10 tend to have better calibration performance when compared to datasets with more labels (such as CIFAR-100 and Tiny-ImageNet) across multiple models. However, it is worth noting that the results on Tiny-ImageNet, which has 200 labels, are generally better than CIFAR-100, despite having more labels. This suggests that the number of labels alone may not be the only factor affecting the calibration performance of a model, and other factors, such as dataset complexity and model architecture, may also play a role. Regarding network architecture, the ResNet-50 is the best calibrated among the four DNNs (ResNet-50, ResNet-110, Wide-ResNet-26-10 and DenseNet-121) tested on both CIFAR-10 and CIFAR-100 datasets. The ResNet-110 performs the worst among the models, especially when trained with focal loss. The Wide-ResNet is generally better calibrated than other DNNs, which may be due to its implicit inner-model ensemble between channels. The Inverse Focal Loss method, however, unlike the results reported in the prior work~\citep{wang2021rethinking}, performs the worst among methods and achieves even higher ECE than the model trained with cross entropy (the Weight Decay column) and the best post-temperature scaling results in only one case, the ResNet-110 on CIFAR-10. This suggests that a loss designed in the opposite direction of this regularization might increase the room for potential improvement by post-hoc calibration, but it makes the model harder to calibrate. Note that we employee the FLSD-53 strategy for focal loss~\citep{mukhoti2020calibrating} to adaptively adjust the gamma sample-wisely, with $\gamma_{focal}=5$ for $\hat{p}\in[0,0.2)$ and $\gamma_{focal}=3$ for $\hat{p}\in[0.2,1)$. We also trained a Global Pooling CNN~\cite{lin2013network} on the 20 Newsgroups~\cite{joachims1996probabilistic}. DFL can obtain better calibration performance in terms of both pre- and post-temperature scaling ECE on NLP tasks. Results of robustness on out-of-distribution datasets can be found in the appendix.

\textbf{Different Metrics} \quad
The methods are evaluated on multiple widely-accepted metrics to evaluate the calibration performance across models. These metrics include Adaptive ECE, Classwise-ECE, and MCE. Adaptive ECE is a metric that measures the expected calibration error of a model, taking into account the distribution of the data. Classwise-ECE is a variant of ECE that measures the calibration error for each class separately. MCE, on the other hand, is a measure of the maximum calibration error over bins. In Figure~\ref{fig:other_metrics}, the results of multiple methods with ResNet-50 and ResNet-110 on the CIFAR-10 dataset are visualized. The figure shows that DFL, is the only method that achieves innately calibrated models and state-of-the-art calibration performance across multiple metrics. Additionally, more results are reported in the appendix of the paper, providing further evidence of the effectiveness of our method for improving the calibration performance.

\begin{figure*}
     \centering
     \begin{subfigure}[b]{0.3\textwidth}
    	\centering
    	\includegraphics[width=\linewidth]{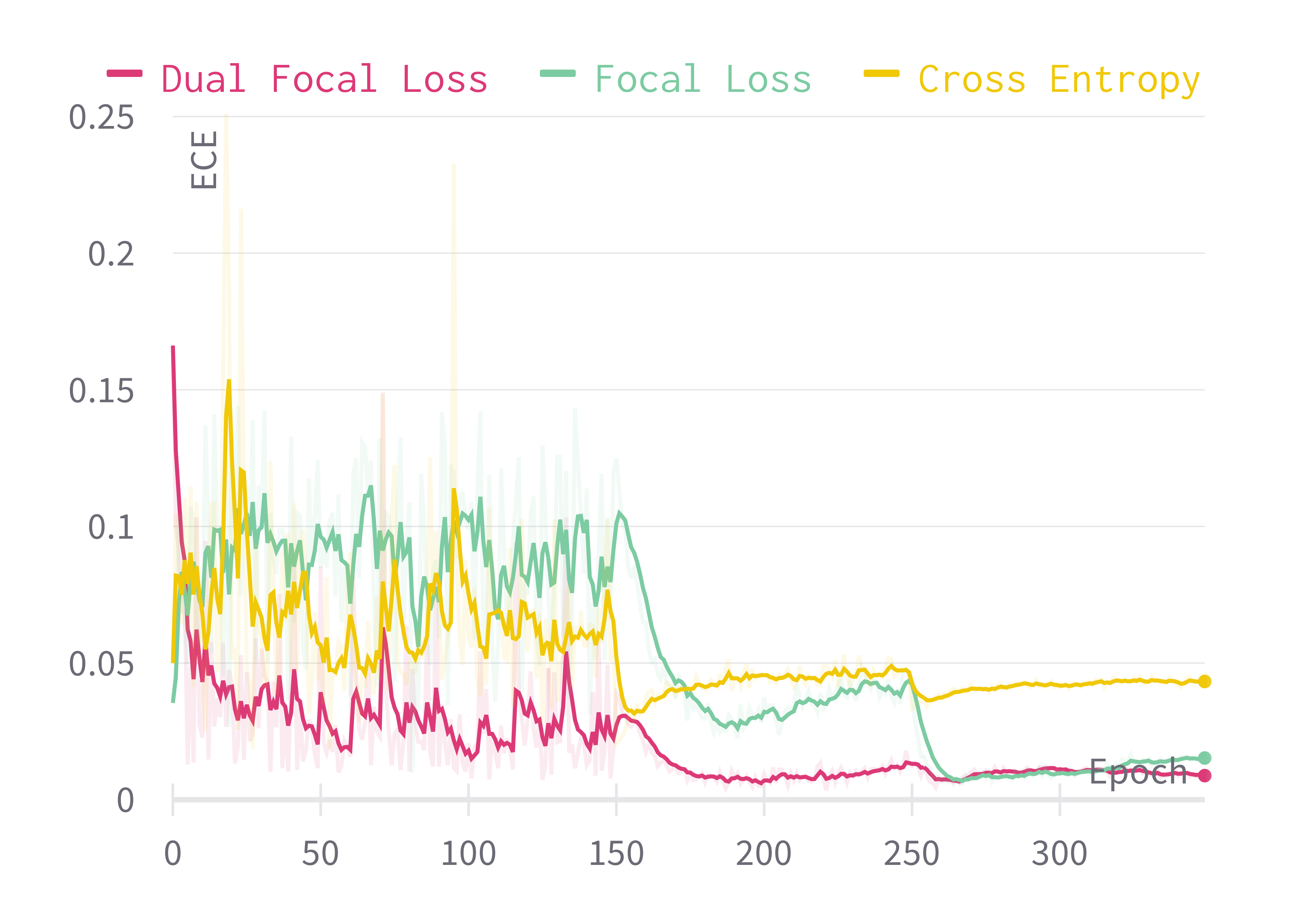}
    	\caption{}
        \label{fig:test_ece_epochs}
     \end{subfigure}
     \hfill
     \begin{subfigure}[b]{0.3\textwidth}
    	\centering
    	\includegraphics[width=\linewidth]{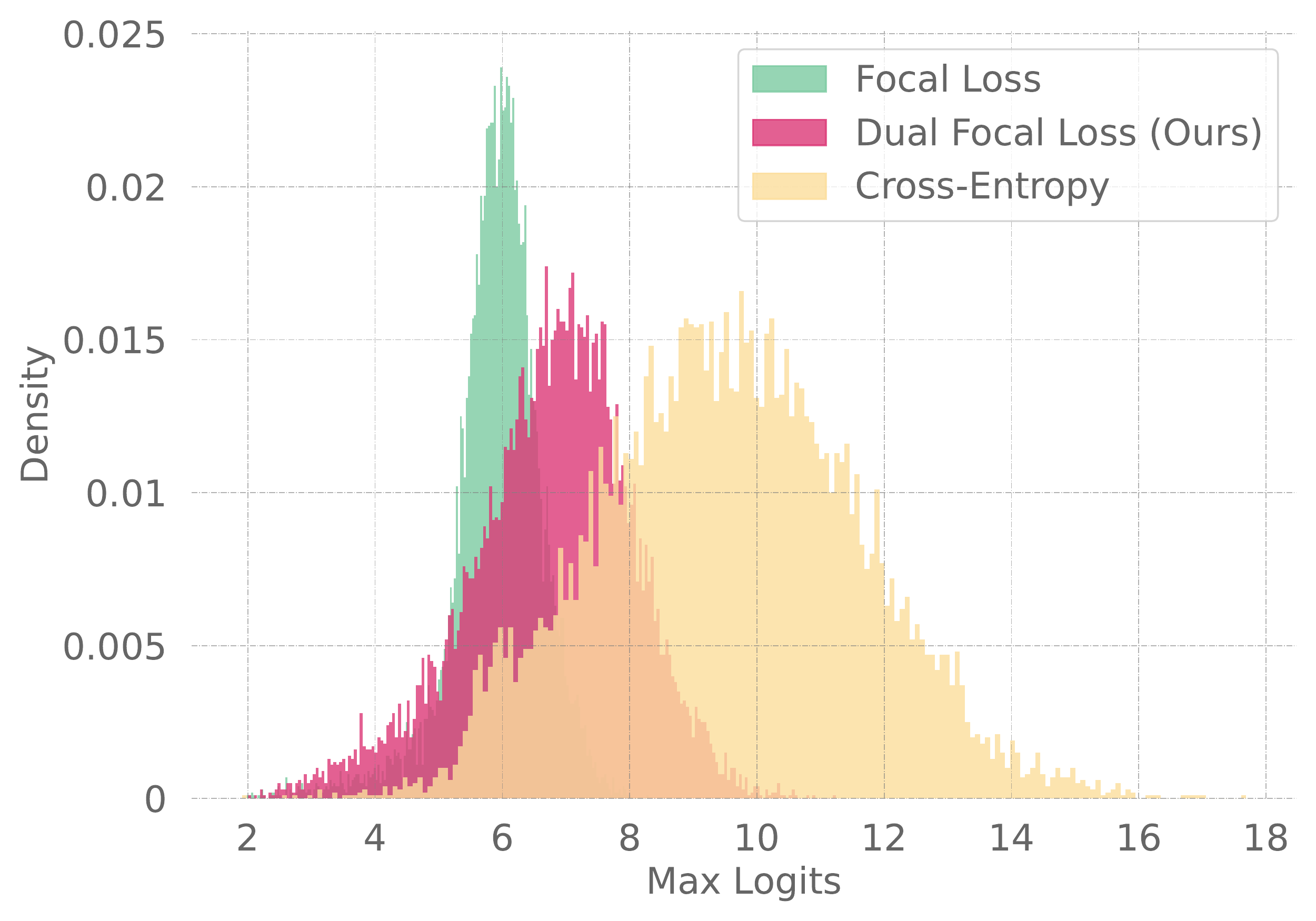}
    	\caption{
    	}
    	\label{fig:maximum logits}
     \end{subfigure}
     \hfill
     \begin{subfigure}[b]{0.3\textwidth}
    	\centering
    	\includegraphics[width=\linewidth]{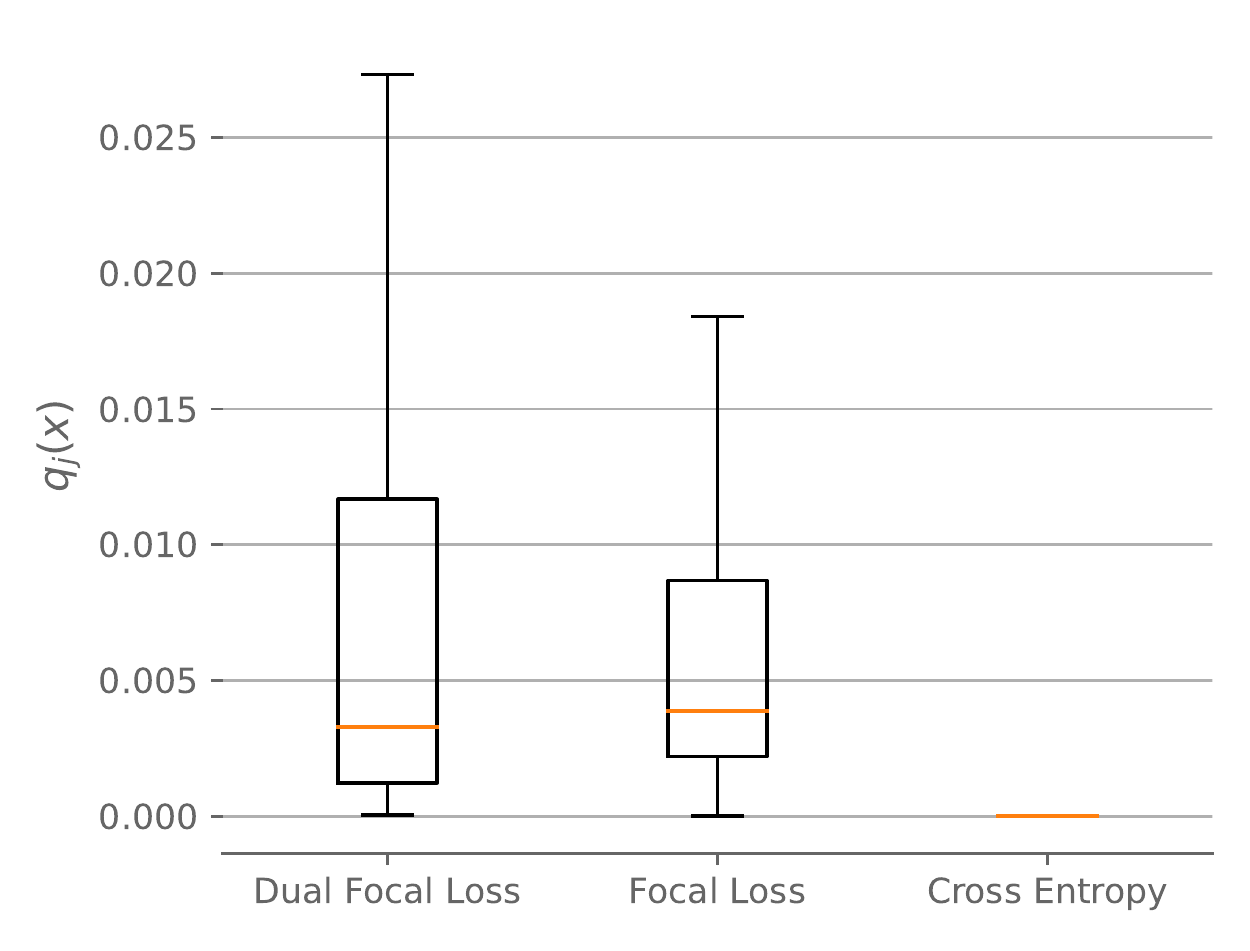}
    	\caption{}
    	\label{fig:distribution_pk_pj}
     \end{subfigure}
        \caption{\textbf{(a) Test ECE with training.} ECE is smoothed by Exponential Moving Average with smoothing factor equals to 0.5; \textbf{(b) Histograms of maximum logits. } All logits are the model outputs before SoftMax layer; \textbf{(c) Boxplot of $q_j(x)$.} The outliers are not shown in this plot. All results are produced by ResNet-50 trained with different methods on CIFAR-10. }
        \label{fig:three graphs}
\end{figure*}

\textbf{Calibration over Training} \quad
Figure~\ref{fig:test_ece_epochs} shows the ECE on the test set for models trained with DFL, FL, and cross-entropy loss over a number of training epochs. The ECE is smoothed using an exponential moving average for better visualization. The figure suggests that after the first few warm-up epochs, where the predicted probability $q_j(x)$ is unstable, the ECE of the DFL-trained models remains at a lower level than the results of models trained using cross-entropy loss and FL. The results shown in Figure~\ref{fig:test_ece_epochs} indicate that when training models with a larger learning rate (0.1 from epoch 1 to 150), FL tends to produce less calibrated models than models trained using cross-entropy loss. On the other hand, models trained with DFL can train calibrated models regardless of the learning rate used. This shows that the DFL is more stable and consistent during the training process than other methods.

\textbf{Reliability Diagram } \quad
Figure~\ref{fig:Reliability Diagram} shows the reliability diagrams for models trained using cross-entropy loss, label smoothing, FL, and DFL on the CIFAR-10 dataset using ResNet-50 architecture.
The figure shows that the DFL retains the best calibration performance over almost all bins. This indicates that the predicted probabilities produced by the model are accurate and reliable across different ranges of predicted probabilities. In contrast, the other methods show a higher deviation from the diagonal line, indicating that the models are less calibrated.

\textbf{Classification Error} \quad
In general, it is often the case that models achieve better calibration performance at the cost of accuracy. Surprisingly, table~\ref{table:error_tab1} presents the classification error on the test set for all methods, and it shows that the DFL improves the classification performance in almost all cases compared to the focal loss. Additionally, the method retains better calibration performance with no trade-off in accuracy on all models trained on CIFAR-100 compared to models trained using cross-entropy loss, which is listed as the weight decay in the table. This suggests that the DFL can improve the calibration of the model without sacrificing its accuracy. This is a significant advantage as it allows for developing accurate and well-calibrated models, which is essential for real-world applications.


\subsection{Comparison with AdaFocal Loss} \label{sec:adafocal}
AdaFocal~\cite{ghosh2022adafocal} proposes an enhanced gamma schedule strategy that selects gamma independently for each training sample based on the model's under/over-confidence on the validation set. On the other hand, DFL focuses on the dual logits to strike a balance between overconfidence and under-confidence, which is independent of AdaFocal. We compared the results under the same training setting and found that DFL and AdaFocal are evenly matched. In fact, we can combine the advantages of both techniques by integrating AdaFocal and DFL, resulting in AdaDualFocal, to achieve improved calibration performance. We conduct experiments on two datasets (cifar10 and cifar100) and four models (Resnet50, Resnet110, Densenet121 and Wide-Resnet), and results for AdaDualFocal demonstrate a consistent enhancement in performance compared to either AdaFocal or DFL. We report the ECE of different methods in Table~\ref{table:adafocal}.
\begin{table}[ht]
\centering
\scriptsize
\resizebox{\linewidth}{!}{%
\begin{tabular}{cccccc}
\toprule
\textbf{Dataset} & \textbf{Method} & \textbf{FLSD-53} & \textbf{AdaFocal} & \textbf{DualFocal} & \textbf{AdaDualFocal}\\

\midrule
\multirow{4}{*}{CIFAR-10}
&Resnet50&1.55	&0.66	&0.46	&\textbf{0.43}\\ 
&Resnet110&1.87	&0.71	&0.98	&\textbf{0.69}\\ 
&Densenet121&1.56	&0.64	&0.81	&\textbf{0.5}\\ 
&WideResnet&1.22	&0.62	&0.57	&\textbf{0.54}\\ 
\midrule
\multirow{4}{*}{CIFAR-100}
&Resnet50&4.5	&1.36	&1.08	&\textbf{1.07}\\
&Resnet110&8.56	&1.4	&2.9	&\textbf{1.14}\\
&Densenet121&3.03	&1.95	&\textbf{1.79}	&1.8\\
&WideResnet&3.73	&1.73	&1.81	&\textbf{1.63}\\

\bottomrule
\end{tabular}}
\caption{\textbf{Comparison with AdaFocal Loss.} }
\vspace{-0.2in}
\label{table:adafocal}
\end{table}

\subsection{Balance between Regularization and Loss of Sample Hardness Information} \label{sec:trade-off}
Although various regularization techniques can be utilized to improve the calibration performance of DNNs, it always leads to the loss of sample hardness, which could limit the potential for calibration improvement \citep{wang2021rethinking}. We mainly attribute it to the sub-optimal solution where the predicted confidences are constrained to a low value to avoid high ECE, which suffers from under-confidence. As analyzed in Sec \ref{sec:theoretical}, our proposed DFL can achieve better trade-offs between $\eta$OC and $\eta$UC. To illustrate it, Figure~\ref{fig:maximum logits} shows the ranges of maximum logit outputs (model outputs before softmax layer) of models as a measure of retained hardness information. As shown in the figure, DFL can be considered as a balance between the cross-entropy and focal loss, which performs regularization while retaining the hardness information of the samples. Figure~\ref{fig:distribution_pk_pj} shows the distribution of the predicted probabilities $q_j(x)$ for different methods. The results shown in Figure~\ref{fig:distribution_pk_pj} indicate that the DFL pushes the $q_j(x)$ to a more extensive range and lower median than the focal loss while also keeping it at a relatively higher value compared to the cross-entropy method. This suggests that the DFL can regularize the model while retaining more of the hardness information of the samples.

\subsection{Ablation Study}\label{sec:abl}
To further explore the effectiveness of the proposed DFL, we conduct ablation study on the variants of $q_j$ in Eq. \ref{eq:DFL}. Our proposed DFL aims at revealing the necessity of the involvement of the dual logit in loss design for calibration and focus on the gap between the confidence logit and dual logit. We acknowledge the existence of various alternatives, such as the involvement of more logits. However, we contend that our approach is theoretically sound and can be extended to scenarios where more logits are involved. Specifically, we can replace C by the mean of different logits in Eq. \ref{eq:DFL}. Thus, Lemma 1 still holds for the scenario with more logits. For completeness, we conduct ablation studies of different logits in dual focal loss including the 2nd and 3rd largest logits lower than the confidence logit, and the mean of different logits. The experiments are conducted with Resnet50 on CIFAR10 and we report the performance in terms of ECE, after temperature scaling ECE(ECE(T)), Adaptive- ECE and Classwise-ECE. As shown in the following table, with the replacement of other dual logits, our method still outperforms the original focal loss~\cite{mukhoti2020calibrating} consistently across different metrics. For the simplicity of theoretical analysis, we only use the maximum logit after the the confidence logit (1st largest in the table) in our proposed DFL.
\begin{table}[ht]
\centering
\scriptsize
\resizebox{\linewidth}{!}{%
\begin{tabular}{cccc}
\toprule
\textbf{Method} & \textbf{Dual Logit} & \textbf{ECE} & \textbf{AdaECE}\\

\midrule
Focal loss & +0.0 & 1.55 & 1.56\\
\midrule
\multirow{2}{*}{Focal Loss with fixed dual logit}
& +0.1 & 1.15 &1.52\\ 
& +0.2 & 2.10 &2.17\\ 
\midrule
\multirow{5}{*}{Dual Focal Loss at other logits}
&+2nd largest logit ranked after $q_{gt}$ & 1.12&1.46\\
&+3rd largest logit ranked after $q_{gt}$ & 1.34&1.35\\
&+mean(1st + 2nd) largest logit ranked after $q_{gt}$ & 0.8&0.68\\
&+mean(1st + 2nd + 3rd) largest logit ranked after $q_{gt}$ & 0.61&0.5\\
&+mean(all other logits lower than $q_{gt}$ & 0.52&0.38\\
\midrule
Dual Focal Loss (Ours)& + largest lower than $q_{gt}$& \bf0.46 & \bf0.66\\ 
\bottomrule
\end{tabular}}
\caption{\textbf{Ablation Study.} Variants of $q_j$ in Dual Focal Loss. Experiments are conducted on the ResNet-50 on CIFAR-10.}
\label{table:abl}
\end{table}


\section{Conclusion}
In conclusion, DFL is a simple and effective method for improving the calibration performance of DNNs. Our method alleviates the under-confidence problem in the focal loss by adding a dual focal term while preserving the hardness information delivered by the output logits. We provide theoretical and practical evidence that DFL has reduced the region size of $\eta$UC and state-of-the-art performance on multiple models and datasets to support the superiority of DFL.
\section*{Acknowledgement}
This work was supported in part by the Australian Research Council under Project DP210101859 and the University of Sydney Research Accelerator (SOAR) Prize. The authors acknowledge the use of the National Computational Infrastructure (NCI) which is supported by the Australian Government, and accessed through the NCI Adapter Scheme and Sydney Informatics Hub HPC Allocation Scheme. The AI training platform supporting this work were provided by High-Flyer AI. (Hangzhou High-Flyer AI Fundamental Research Co., Ltd.)

\bibliography{example_paper}
\bibliographystyle{icml2023}

\newpage
\appendix
\onecolumn
\section{Proof of Theorem 1}

Note that $\phi(v_i) = (1-v_i+f(v_i))^{\gamma} + (\frac{\partial f}{\partial v_i} - 1)\gamma(1-v_i+f(v_i))^{\gamma-1}v_i \text{ log }v_i$ and $\eta_i$ can be formulated as $\frac{\frac{q^*_i}{\phi(q^*_i)}}{\sum^K_k \frac{q^*_k}{\phi(q^*_k)}}$ in Eq. \ref{eq:eta_2}. Now we simplify Eq. \ref{eq:eta_2} through taking $h(v_i) = \frac{v_i}{\phi(v_i)}$ and function $h(v)$ can be formulated as
\begin{equation}
h(v) = \frac{v}{\phi(v)} = 
\begin{cases}
    \frac{v}{1-\gamma v \text{ log }v} , & i = j\\
    \frac{v}{(1-v+C)^{\gamma}  - \gamma(1-v+C)^{\gamma-1}v \text{ log }v}, & i\neq j\\
\end{cases}
\end{equation}
With function $h$, Eq. \ref{eq:eta_2} can be rewritten as
\begin{equation}
    \eta_i = \frac{h(q^*_i)}{\sum^K_k h(q^*_k)}.
\end{equation}
Considering the scenario $i=j$, we take the derivative of $h(v)$ as
\begin{equation}
\begin{aligned}
    \frac{\partial h}{\partial v} &  = \frac{1}{1-\gamma v \text{ log }v} - \frac{v(-\gamma \text{ log }v -\gamma)}{(1-\gamma v \text{ log }v)^2}\\
    & = \frac{1-\gamma v \text{ log }v + \gamma v \text{ log }v + \gamma v}{(1-\gamma v \text{ log }v)^2}\\
    & = \frac{1+\gamma v}{(1-\gamma v \text{ log }v)^2} > 0\\
\end{aligned}
\end{equation}
Considering the scenario $i \neq j$, we take the derivative of $h(v)$ as
\begin{equation}
\begin{aligned}
    \frac{\partial h}{\partial v} &  = \frac{(1-v+C)^{\gamma}  - \gamma(1-v+C)^{\gamma-1}v \text{ log }v + 2\gamma v (1-v+C)^{\gamma-1} + \gamma v (1-v+C)^{\gamma-1} \text{ log }v }{[(1-v+C)^{\gamma} - \gamma(1-v+C)^{\gamma-1}v \text{ log }v]^2}\\
    & \;\;\; - \frac{(\gamma-1)\gamma v^2 (1-v+C)^{\gamma-2} \text{ log }v}{[(1-v+C)^{\gamma} - \gamma(1-v+C)^{\gamma-1}v \text{ log }v]^2}\\
    & = \frac{(1-v+C)^{\gamma} + 2\gamma v (1-v+C)^{\gamma-1} - (\gamma-1)\gamma v^2 (1-v+C)^{\gamma-2} \text{ log }v}{[(1-v+C)^{\gamma} - \gamma(1-v+C)^{\gamma-1}v \text{ log }v]^2}\\
\end{aligned}
\end{equation}
If $\gamma \geq 1$, it is easy to see that $\frac{\partial h}{\partial v} > 0$. If $\gamma \in [0,1)$, we denote the denominator as $t(v)$, which can be formulated as.
\begin{equation}
\begin{aligned}
    t(v) & = (1-v+C)^{\gamma} + 2\gamma v (1-v+C)^{\gamma-1} - (\gamma-1)\gamma v^2 (1-v+C)^{\gamma-2} \text{ log }v\\
    & = (1-v+C)^{\gamma-2} \Big[(1-v+C)^2 + 2 \gamma v (1-v+C) - \gamma^2 v^2 \text{ log }v + \gamma v^2 \text{ log }v \Big]\\
\end{aligned}
\end{equation}
Since $(1-v+C)^{\gamma-2} \geq 0$ and $- \gamma^2 v^2 \text{ log }v \geq 0$, we use function $u(v)$ to cover the remaining terms as
\begin{equation}
    u(v) = (1-v+C)^2 + 2 \gamma v (1-v+C) + \gamma v^2 \text{ log }v,
\end{equation}
To discover the sign of $u(v)$, we compute the derivative and $u(1)$ as
\begin{equation}
\begin{aligned}
    & u(1) = C^2 + 2 \gamma C \geq 0,\\
    & \frac{\partial u }{\partial v} = -2(1-v+C) + 2\gamma(1-v+C) - 2\gamma v + 2\gamma v \text{ log }v + \gamma v\\
    & \;\;\;\;\; = 2v -2 -2C + 2\gamma + 2\gamma C -3\gamma v + 2\gamma v \text{ log }v\\
\end{aligned}
\end{equation}
Taking the advantage of the fact that
\begin{equation}
\begin{aligned}
    & \frac{\partial u }{\partial v}|_{v=0} = 2(\gamma - 1) + 2C(\gamma-1) < 0,\\
    & \frac{\partial u }{\partial v}|_{v=1} = -\gamma < 0,\\
    & \frac{\partial^2 u}{\partial v^2} = 2 - \gamma + 2\gamma \text{ log }v,\\
    & \frac{\partial^3 u}{\partial v^3} = \frac{2\gamma}{v} > 0,\\
\end{aligned}
\end{equation}
we can see $\frac{\partial u }{\partial v}$ is convex, which makes $\frac{\partial u }{\partial v} < 0$ for all the $v \in (0,1)$. Thus, $u(v)$ is a decreasing function. Together with the fact that $u(1) > 0$, $u(v) > 0$ all the $v \in (0,1)$. Similarly, we can conclude that $\frac{\partial h}{\partial v} > 0$ in all the scenarios, which makes $h$ a strictly increasing function. In other words, $q^*_a < q^*_b \Rightarrow h(q^*_a) < h(q^*_b)$. Since we have $\eta_i = \frac{h(q^*_i)}{\sum^K_k h(q^*_k)}$, it is easy to see that $q^*_a < q^*_b \Rightarrow \eta_a < \eta_b$. Thus, our proposed dual focal loss is strictly order-preserving, which is sufficient for classification-calibration \citep{zhang2004statistical}.

\section{Proof of Lemma 1}
We explore the property of function $\phi$ via the its derivative as
\begin{equation} \label{eq:phi_partial}
\begin{aligned}
    \frac{\partial \phi}{\partial v_i} & = (-1+\frac{\partial f}{\partial v_i})\gamma(1-v_i+f(v_i))^{\gamma-1} + \frac{\partial^2 f}{\partial v_i^2} \gamma(1-v_i+f(v_i))^{\gamma-1}v_i\text{ log }v_i\\
    & \;\;\; + (-1+\frac{\partial f}{\partial v_i})(\frac{\partial f}{\partial v_i} - 1) \gamma(\gamma-1) (1-v_i+f(v_i))^{\gamma-2} v_i\text{ log }v_i \\
    & \;\;\; + (\frac{\partial f}{\partial v_i} - 1)\gamma(1-v_i+f(v_i))^{\gamma-1}\text{ log }v_i + (\frac{\partial f}{\partial v_i} - 1)\gamma(1-v_i+f(v_i))^{\gamma-1}\\
    & = \gamma(1-v_i+f(v_i))^{\gamma-2} \Big( 2(\frac{\partial f}{\partial v_i}-1) (1-v_i+f(v_i)) +  \frac{\partial^2 f}{\partial v_i^2} (1-v_i+f(v_i)) v_i \text{ log }v_i \\
    & \;\;\; + (\frac{\partial f}{\partial v_i}-1)^2 (\gamma-1) v_i \text{ log }v_i + (\frac{\partial f}{\partial v_i}-1)(1-v_i+f(v_i))\text{ log }v_i \Big)\\
\end{aligned}
\end{equation}
According to the definition in Eq. \ref{eq:DFL}, for function $f$, it maps $v_i$ to the second maximum value $v_j$ in the ascending order $v_{1:K}$ after $v_{gt}$. Formally, the function $f$ is defined as
\begin{equation}
\begin{aligned}
    & f(v_i) = v_j = C\\
    & \text{where } v_j = \max_i \{v_i|v_i < v_{gt}\}\\
\end{aligned}
\end{equation}
Thus, if $i=j$, $\frac{\partial f}{\partial v_i} = 1$, otherwise, $\frac{\partial f}{\partial v_i} = 0$. Eq. \ref{eq:phi_partial} can be simplified as
\begin{equation}
    \frac{\partial \phi}{\partial v_i} =
    \begin{cases}
        0, & i = j\\
        \gamma(1-v_i+C)^{\gamma-2} \Big( -2 (1-v_i+C) + (\gamma-1)v_i \text{ log }v_i - (1-v_i+C) \text{ log }v_i \Big), & i \neq j \\
    \end{cases}
\end{equation}
Now we consider the scenario where $i\neq j$
\begin{equation}
\begin{aligned}
    \frac{\partial \phi}{\partial v_i} &  = \gamma(1-v_i+C)^{\gamma-2} \Big( 2(v_i-1-C)+\gamma v_i \text{ log }v_i - v_i \text{ log }v_i - (1+C) \text{ log }v_i +   v_i \text{ log }v_i \Big)\\
    & = \gamma(1-v_i+C)^{\gamma-2} \Big( 2(v_i-1-C) + \gamma v_i \text{ log }v_i - (1+C) \text{ log }v_i \Big)\\
\end{aligned}
\end{equation}
Since $ \gamma(1-v_i+C)^{\gamma-2} > 0$, we let $s(v) = 2(v-1-C) + \gamma v \text{ log }v - (1+C) \text{ log }v$ and explore its proprieties as
\begin{equation}
\begin{aligned}
    & \frac{\partial s}{\partial v} = 2 + \gamma logv + \gamma - \frac{1+C}{v},\\
    & \frac{\partial^2 s}{\partial v^2} = \frac{\gamma}{v} + \frac{1+C}{v^2}\\
    & s(0) = \infty, \; s(1) = -2C, \; \frac{\partial s}{\partial v}|_{v=1} = 1+\gamma-C\\
\end{aligned}
\end{equation}
Since $\frac{\gamma}{v} + \frac{1+C}{v^2} > 0$ for all $v$, $s(v)$ is convex. Furthermore, according to the definition of $v_j$, $0 \leq C < 1$. Thus, $s(1) \leq 0$ and $\frac{\partial s}{\partial v}|_{v=1} > 0$. Taking the advantage of intermediate value theorem, there exist $v' \in (0,1)$ such that $s(v)' = 0$ and $v'$ is unique since $s(v)$ is convex. Thus, $\frac{\partial \phi}{\partial v_i} = 0$ at $v'_i$,  $\frac{\partial \phi}{\partial v_i} > 0$ for $v \in [0,v'_i)$ and $\frac{\partial \phi}{\partial v_i} < 0$ for $v \in (v'_i,1]$.

\section{Dataset Description}
\label{dataset}
We evaluate the performance of our proposed Dual Focal Loss on multiple datasets, including CIFAR-10/100~\cite{krizhevsky2009learning} and Tiny-ImageNet~\cite{deng2009imagenet}, to assess its calibration performance. Additionally, we include evaluations of robustness to Out-of-Distribution (OoD) data using datasets such as SVHN~\cite{goodfellow2013multi} and CIFAR-10-C~\cite{hendrycks2018benchmarking}. Below are the specific details of each dataset used in our evaluations:

\textbf{CIFAR-10/100}: The CIFAR-10 dataset includes $60,000$ 32x32 color images, divided into 10 classes with $6,000$ images per class. There are $50,000$ training images and $10,000$ test images. The CIFAR-100 dataset has 100 classes and 600 images per class. For CIFAR-10/100, we use $5,000$ images from the training set for validation.

\textbf{Tiny ImageNet}: This dataset is a subset of ImageNet from the Large Scale Visual Recognition Challenge (ILSVRC) and contains $100,000$ colored images of 200 classes, with 500 images per class. Each image is downsized to 64x64 pixels.

\noindent
\textbf{SVHN}: The Street View House Numbers (SVHN) dataset is a real-world image dataset designed for developing machine learning and object recognition algorithms. It contains over 600,000 images of house numbers taken from Google Street View, each belonging to one of 10 classes. We evaluate the performance of the methods on this dataset by assessing its ability to handle dataset shift, specifically by testing its performance on the testing set.

\noindent
\textbf{CIFAR-10-C}: The CIFAR-10-C dataset is a version of the CIFAR-10 dataset that includes images corrupted with various types of noise. The first 10,000 images in this dataset are test set images corrupted at severity level 1, while the last 10,000 images are test set images corrupted at severity level 5. In our experiments, we use the Gaussian Noise corruption with a severity level of 5 to evaluate the robustness of the methods.

\section{Comparison Methods}
\label{methods}
To evaluate the effectiveness of our proposed algorithm, we include several comparison methods in our experiments. The details of these comparison methods are provided below:

\noindent
\textbf{Weight Decay}: We train the model with the Cross-Entropy loss and multiple weight decay values, including $1\times10^{-4}$, $5\times10^{-4}$, and $1\times10^{-3}$. We report the result of $5\times10^{-4}$, which has the best calibration performance.

\noindent
\textbf{Brier Loss}~\cite{brier1950verification}: This is a square error loss calculated between the softmax logits and the one-hot labels.

\noindent
\textbf{MMCE Loss}~\cite{kumar2018trainable}: MMCE is a RKHS kernel-based trainable auxiliary loss used alongside the NLL loss to improve calibration performance.

\noindent
\textbf{Label Smoothing}\cite{szegedy2016rethinking}: Label smoothing replaces the one-hot encoded label vector with a mixture of labels and the uniform distribution. We follow the settings in\cite{mukhoti2020calibrating} and set the smoothing vector used in this work to $0.05$.

\noindent
\textbf{Inverse Focal Loss}~\cite{wang2021rethinking}: The Inverse Focal Loss is an inverted version of the standard focal loss, which aims to maximize the potential room for post-hoc calibration.

\noindent
\textbf{Focal Loss}~\cite{mukhoti2020calibrating}: FLSD-53 is a simplification of the sample-dependent $\gamma$ approach. Mukhoti et al. \cite{mukhoti2020calibrating} introduce a schedule mechanism instead of the original fixed one. In particular, with $\gamma_{focal}=5$ for $\hat{p}\in[0,0.2)$ and $\gamma_{focal}=3$ for $\hat{p}\in[0.2,1)$.

\section{Performance on Different Metrics and Robustness on Dataset Shift}
Adaptive-ECE is a measure of calibration performance that addresses the bias of equal-width binning scheme of ECE. It adapts the bin-size to the number of samples and ensures that each bin is evenly distributed with samples. The formula for Adaptive-ECE is as follows:

\begin{equation}
\text{Adaptive-ECE}=\sum_{i=1}^{\mathbb{B}}\frac{|B_i|}{N} \left| I_i - C_i \right| \text{ s.t. } \forall i,j\cdot|B_i|=|B_j|
\end{equation}

Table \ref{table:ada_ece_tab1} shows that Adaptive-ECE has competitive performance with state-of-the-art results in almost all cases, with only a $0.1\%$ gap between Focal Loss on Tiny-ImageNet. Classwise-ECE is another measure of calibration performance that addresses the deficiency of ECE in only measuring the calibration performance of the single predicted class. It can be formulated as:
\begin{equation}
\text{Classwise-ECE}=\frac{1}{\mathcal{K}} \sum_{i=1}^{\mathbb{B}} \sum_{j=1}^{\mathcal{K}} \frac{|B_{i,j}|}{N} \left|I_{i,j} - C_{i,j} \right|
\end{equation}
where $B_{i,j}$ denotes the set of samples with the $j^{\textrm{th}}$ class label in the $i^{\textrm{th}}$ bin, $I_{i,j}$ and $C_{i,j}$ represents the accuracy and confidence of samples in $B_{i,j}$. 

Table \ref{table:classece_tab1} demonstrates the superior performance of Classwise-ECE, as it shows that DFL not only calibrates the confidence of the predicted label, but also the probabilities of all other class labels. Table \ref{table:nll_tab1} displays competitive NLL results, with DFL achieving a better NLL on more complex datasets such as CIFAR-100 and Tiny-ImageNet. This indicates that DFL acts as a regularization method, which can also be applied to other tasks related to overfitting, such as improving model robustness. Table \ref{table:mce_tab1} illustrates the competitive results of MCE, with DFL achieving the state-of-the-art results in most cases, particularly before temperature scaling. DFL outperforms other methods significantly. Table \ref{table:auroc_tab1} shows the AUROC score measured on different methods. DFL can achieve the competitive results in most cases.
\begin{table*}[!t]
	\centering
	\scriptsize
	\resizebox{\linewidth}{!}{%
		\begin{tabular}{cccccccccccccccc}
			\toprule
			\textbf{Dataset} & \textbf{Model} & \multicolumn{2}{c}{\textbf{Weight Decay}} &
			\multicolumn{2}{c}{\textbf{Brier Loss}} & \multicolumn{2}{c}{\textbf{MMCE}} &
			\multicolumn{2}{c}{\textbf{Label Smoothing}} & \multicolumn{2}{c}{\textbf{Inverse Focal Loss}} &
			\multicolumn{2}{c}{\textbf{Focal Loss}} &
			\multicolumn{2}{c}{\textbf{Dual Focal}} \\
			\textbf{} & \textbf{} &
			\multicolumn{2}{c}{\citep{guo2017calibration}} &
			\multicolumn{2}{c}{\citep{brier1950verification}} & \multicolumn{2}{c}{\citep{kumar2018trainable}} &
			\multicolumn{2}{c}{\citep{szegedy2016rethinking}} & \multicolumn{2}{c}{\citep{wang2021rethinking}} &
			\multicolumn{2}{c}{\citep{mukhoti2020calibrating}} &
			\multicolumn{2}{c}{Ours} \\
			&& Pre T & Post T & Pre T & Post T & Pre T & Post T & Pre T & Post T & Pre T & Post T & Pre T & Post T & Pre T & Post T \\
			\midrule
			\multirow{4}{*}{CIFAR-100} & ResNet-50&17.52&3.42(2.1)&6.52&3.64(1.1)&15.32&2.38(1.8)&7.81&4.01(1.1)&17.8&3.70(2.3)&4.5&2.0(1.1)&\textbf{1.23}&\textbf{1.23(1.0)}\\
			& ResNet-110&19.05&5.86(2.3)&7.73&4.53(1.2)&19.14&4.85(2.3)&11.12&8.59(1.1)&19.4&6.35(2.6)&8.55&3.96(1.2)&\textbf{3.16}&\textbf{3.16(1.0)}\\
			& Wide-ResNet-26-10&15.33&2.89(2.2)&4.22&2.81(1.1)&13.16&4.25(1.9)&5.1&5.1(1.0)&16.9&2.29(2.5)&2.75&\textbf{1.6(1.1)}&\textbf{2.03}&2.03(1.0)\\
			& DenseNet-121&20.98&5.09(2.3)&5.04&2.56(1.1)&19.13&3.07(2.1)&12.83&8.92(1.2)&19.4&2.82(2.3)&3.55&\textbf{1.24(1.1)}&\textbf{1.63}&1.63(1.0)\\
			\midrule
			\multirow{4}{*}{CIFAR-10} & ResNet-50&4.33&2.14(2.5)&1.74&1.23(1.1)&4.55&2.16(2.6)&3.89&2.92(0.9)&4.38&2.15(1.5)&1.56&1.26(1.1)&\textbf{0.66}&\textbf{0.66(1.0)}\\
			& ResNet-110&4.4&1.99(2.8)&2.6&1.7(1.2)&5.06&2.52(2.8)&4.44&4.44(1.0)&4.33&2.66(2.9)&2.07&1.67(1.1)&\textbf{1.23}&\textbf{1.23(1.0)}\\
			& Wide-ResNet-26-10&3.23&1.69(2.2)&1.7&1.7(1.0)&3.29&1.6(2.2)&4.27&2.44(0.8)&3.67&2.06(2.7)&1.52&\textbf{1.38(0.9)}&\textbf{1.42}&1.42(1.0)\\
			& DenseNet-121&4.51&2.13(2.4)&2.03&2.03(1.0)&5.1&2.29(2.5)&4.42&3.33(0.9)&4.61&2.65(2.8)&1.42&1.42(1.0)&\textbf{0.80}&\textbf{0.80(1.0)}\\
			\midrule
			Tiny-ImageNet & ResNet-50&15.23&5.41(1.4)&4.37&4.07(0.9)&13.0&5.56(1.3)&15.28&6.29(0.7)&11.4&6.37(1.3)&\textbf{1.42}&\textbf{1.42(1.0)}&1.52&1.52(1.0)\\
			\bottomrule
		\end{tabular}%
	}
	
	\caption{\textbf{Adaptive ECE $(\%)$ evaluated for different methods.}  Both pre and post temperature scaling results are reported. Optimal temperature is included in brackets.  (calculated temperature on best ECE) }
	\label{table:ada_ece_tab1}
\end{table*}

\begin{table*}[!t]
	\centering
	\scriptsize
	\resizebox{\linewidth}{!}{%
		\begin{tabular}{cccccccccccccccc}
			\toprule
			\textbf{Dataset} & \textbf{Model} & \multicolumn{2}{c}{\textbf{Weight Decay}} &
			\multicolumn{2}{c}{\textbf{Brier Loss}} & \multicolumn{2}{c}{\textbf{MMCE}} &
			\multicolumn{2}{c}{\textbf{Label Smoothing}} & \multicolumn{2}{c}{\textbf{Inverse Focal Loss}} &
			\multicolumn{2}{c}{\textbf{Focal Loss}} &
			\multicolumn{2}{c}{\textbf{Dual Focal}} \\
			\textbf{} & \textbf{} &
			\multicolumn{2}{c}{\citep{guo2017calibration}} &
			\multicolumn{2}{c}{\citep{brier1950verification}} & \multicolumn{2}{c}{\citep{kumar2018trainable}} &
			\multicolumn{2}{c}{\citep{szegedy2016rethinking}} & \multicolumn{2}{c}{\citep{wang2021rethinking}} &
			\multicolumn{2}{c}{\citep{mukhoti2020calibrating}} &
			\multicolumn{2}{c}{Ours} \\
			&& Pre T & Post T & Pre T & Post T & Pre T & Post T & Pre T & Post T & Pre T & Post T & Pre T & Post T  & Pre T & Post T \\
			\midrule
			
			\multirow{4}{*}{CIFAR-100} & ResNet-50 & 0.38 & 0.22(2.1)&0.22&0.20(1.1)&0.34&0.21(1.8)&0.23&0.21(1.1)&0.38&0.20(2.3)&0.20&0.20(1.1)&\textbf{0.19}&\textbf{0.19(1.0)}\\
			& ResNet-110&0.41&0.21(2.3)&0.24&0.23(1.2)&0.42&0.22(2.3)&0.26&0.22(1.1)&0.41&0.21(2.6)&0.24&0.21(1.2)&\textbf{0.21}&\textbf{0.21(1.0)}\\
			& Wide-ResNet-26-10&0.34&0.20(2.2)&0.19&0.19(1.1)&0.31&0.20(1.9)&0.21&0.21(1.0)&0.36&0.21(2.5)&\textbf{0.18}&0.19(1.1)&0.19&\textbf{0.19(1.0)}\\
			& DenseNet-121&0.45&0.23(2.3)&0.20&0.21(1.1)&0.42&0.24(2.1)&0.29&0.24(1.2)&0.41&0.24(2.3)&0.19&0.20(1.1)&\textbf{0.20}&\textbf{0.20(1.0)}\\
			\midrule
			\multirow{4}{*}{CIFAR-10} & ResNet-50&0.91&0.45(2.5)&0.46&0.42(1.1)&0.94&0.52(2.6)&0.71&0.51(0.9)&0.91&0.44(2.8)&0.42&0.42(1.1)&\textbf{0.35}&\textbf{0.35(1.0)}\\
			& ResNet-110&0.91&0.50(2.8)&0.59&0.50(1.2)&1.04&0.55(2.8)&0.66&0.66(1.0)&0.89&0.45(2.9)&0.48&0.44(1.1)&\textbf{0.35}&\textbf{0.35(1.0)}\\
			& Wide-ResNet-26-10&0.68&0.37(2.2)&0.44&0.44(1.0)&0.70&0.35(2.2)&0.80&0.45(0.8)&0.77&0.39(2.7)&0.41&\textbf{0.31(0.9)}&\textbf{0.34}&0.34(1.0)\\
			& DenseNet-121&0.92&0.47(2.4)&0.46&0.46(1.0)&1.04&0.57(2.5)&0.60&0.50(0.9)&0.94&0.51(2.8)&0.41&0.41(1.0)&\textbf{0.35}&\textbf{0.35(1.0)}\\
			\midrule
			Tiny-ImageNet & ResNet-50&0.22&0.16(1.4)&0.16&0.16(0.9)&0.21&0.16(1.3)&0.21&0.17(0.7)&0.16&\textbf{0.14(1.3)}&0.16&0.16(1.0)&\textbf{0.16}&0.16(1.0)\\

			\bottomrule
		\end{tabular}%
	}
	
	\caption{\textbf{Classwise-ECE $(\%)$ evaluated for different methods.}  Both pre and post temperature scaling results are reported. Optimal temperature is included in brackets.  (calculated temperature on best ECE) }
	\label{table:classece_tab1}

\end{table*}

\begin{table*}[!t]
	\centering
	\scriptsize
	\resizebox{\linewidth}{!}{%
		\begin{tabular}{cccccccccccccccc}
			\toprule
			\textbf{Dataset} & \textbf{Model} & \multicolumn{2}{c}{\textbf{Weight Decay}} &
			\multicolumn{2}{c}{\textbf{Brier Loss}} & \multicolumn{2}{c}{\textbf{MMCE}} &
			\multicolumn{2}{c}{\textbf{Label Smoothing}} & \multicolumn{2}{c}{\textbf{Inverse Focal Loss}} &
			\multicolumn{2}{c}{\textbf{Focal Loss}} &
			\multicolumn{2}{c}{\textbf{Dual Focal}} \\
			\textbf{} & \textbf{} &
			\multicolumn{2}{c}{\citep{guo2017calibration}} &
			\multicolumn{2}{c}{\citep{brier1950verification}} & \multicolumn{2}{c}{\citep{kumar2018trainable}} &
			\multicolumn{2}{c}{\citep{szegedy2016rethinking}} & \multicolumn{2}{c}{\citep{wang2021rethinking}} &
			\multicolumn{2}{c}{\citep{mukhoti2020calibrating}} &
			\multicolumn{2}{c}{Ours} \\ \\
			&& Pre T & Post T & Pre T & Post T & Pre T & Post T & Pre T & Post T & Pre T & Post T & Pre T & Post T  & Pre T & Post T \\
			\midrule
			
			\multirow{4}{*}{CIFAR-100} & ResNet-50
			&44.34&12.75(2.1)&36.75&21.61(1.1)&39.53&11.99(1.8)&26.11&18.58(1.1)&50.22&16.20(2.3)&16.12&27.18(1.1)&\textbf{5.29}&\textbf{5.29(1.0)}\\
			& ResNet-110&55.92&22.65(2.3)&24.85&13.56(1.2)&50.69&19.23(2.3)&36.23&30.46(1.1)&53.59&19.68(2.6)&22.57&10.94(1.2)&\textbf{8.10}&\textbf{8.10(1.0)}\\
			& Wide-ResNet-26-10&49.36&14.18(2.2)&14.68&13.42(1.1)&40.13&16.5(1.9)&23.79&23.79(1.0)&52.90&12.50(2.5)&\textbf{10.17}&\textbf{9.73(1.1)}&11.89&11.89(1.0)\\
			& DenseNet-121&56.28&21.63(2.3)&15.47&8.55(1.1)&49.97&13.02(2.1)&43.59&29.95(1.2)&53.11&8.45(2.3)&\textbf{9.68}&\textbf{5.68(1.1)}&10.14&10.14(1.0)\\
			\midrule
			\multirow{4}{*}{CIFAR-10} &ResNet-50&36.65&20.6(2.5)&31.54&22.46(1.1)&60.06&23.6(2.6)&35.61&40.51(0.9)&49.74&34.1(2.8)&\textbf{14.89}&26.37(1.1)&24.24&\textbf{24.24(1.0)}\\
			& ResNet-110&44.25&29.98(2.8)&25.18&22.73(1.2)&67.52&31.87(2.8)&45.72&45.72(1.0)&39.81&32.44(2.9)&18.95&17.35(1.1)&\textbf{12.59}&\textbf{12.59(1.0)}\\
			& Wide-ResNet-26-10&48.17&26.63(2.2)&77.15&77.15(1.0)&36.82&32.33(2.2)&24.89&37.53(0.8)&33.51&74.52(2.7)&74.07&36.56(0.9)&\textbf{26.27}&\textbf{26.27(1.0)}\\
			& DenseNet-121&45.19&32.52(2.4)&19.39&19.39(1.0)&43.29&27.03(2.5)&45.5&53.57(0.9)&52.11&33.79(2.8)&\textbf{13.36}&\textbf{13.36(1.0)}&52.11&52.11(1.0)\\
			\midrule
			Tiny-ImageNet & ResNet-50&30.83&13.33(1.4)&8.41&12.82(0.9)&34.72&12.52(1.3)&25.48&17.2(0.7)&30.13&11.53(1.3)&\textbf{3.76}&\textbf{3.76(1.0)}&4.82&4.82(1.0)\\
			\bottomrule
		\end{tabular}%
	}
	
	\caption{\textbf{MCE $(\%)$ evaluated for different methods.}  Both pre and post temperature scaling results are reported. Optimal temperature is included in brackets (calculated temperature on best ECE). }
	\label{table:mce_tab1}

\end{table*}

\begin{table*}[!t]
	\centering
	\scriptsize
	\resizebox{\linewidth}{!}{%
		\begin{tabular}{cccccccccccccccc}
			\toprule
			\textbf{Dataset} & \textbf{Model} & \multicolumn{2}{c}{\textbf{Weight Decay}} &
			\multicolumn{2}{c}{\textbf{Brier Loss}} & \multicolumn{2}{c}{\textbf{MMCE}} &
			\multicolumn{2}{c}{\textbf{Label Smoothing}} & \multicolumn{2}{c}{\textbf{Inverse Focal Loss}} &
			\multicolumn{2}{c}{\textbf{Focal Loss}} &
			\multicolumn{2}{c}{\textbf{Dual Focal}} \\
			\textbf{} & \textbf{} &
			\multicolumn{2}{c}{\citep{guo2017calibration}} &
			\multicolumn{2}{c}{\citep{brier1950verification}} & \multicolumn{2}{c}{\citep{kumar2018trainable}} &
			\multicolumn{2}{c}{\citep{szegedy2016rethinking}} & \multicolumn{2}{c}{\citep{wang2021rethinking}} &
			\multicolumn{2}{c}{\citep{mukhoti2020calibrating}} &
			\multicolumn{2}{c}{Ours} \\ \\
			&& Pre T & Post T & Pre T & Post T & Pre T & Post T & Pre T & Post T & Pre T & Post T & Pre T & Post T  & Pre T & Post T \\
			\midrule
			
			\multirow{4}{*}{CIFAR-100} & ResNet-50
			&153.67&106.83(2.1)&99.63&99.57(1.1)&125.28&101.92(1.8)&121.02&120.19(1.1)&170.9&104.8(2.3)&88.03&88.27(1.1)&\textbf{87.75}&\textbf{87.75(1.0)}\\
			& ResNet-110&179.21&104.63(2.3)&110.72&111.81(1.2)&180.54&106.73(2.3)&133.11&129.76(1.1)&210.3&110.8(2.6)&89.92&88.93(1.2)&\textbf{88.81}&\textbf{88.81(1.0)}\\
			& Wide-ResNet-26-10&140.1&91.0(2.2)&84.62&85.77(1.1)&119.58&95.92(1.9)&108.06&108.06(1.0)&173&100.6(2.5)&\textbf{76.92}&\textbf{78.14(1.1)}&78.67&78.67(1.0)\\
			& DenseNet-121&205.61&119.23(2.3)&98.31&98.74(1.1)&166.65&113.24(2.1)&142.04&136.28(1.2)&178.6&115.8(2.3)&\textbf{85.47}&86.06(1.1)&85.82&\textbf{85.82(1.0)}\\
			\midrule
			\multirow{4}{*}{CIFAR-10} &ResNet-50&41.21&20.38(2.5)&18.36&18.36(1.1)&44.83&21.58(2.6)&27.68&27.69(0.9)&48.3&21.0(2.8)&17.55&17.37(1.1)&\textbf{17.02}&\textbf{17.02(1.0)}\\
			& ResNet-110&47.51&21.52(2.8)&20.44&19.60(1.2)&55.71&24.61(2.8)&29.88&29.88(1.0)&52.9&22.3(2.9)&18.54&18.24(1.1)&\textbf{17.98}&\textbf{17.98(1.0)}\\
			& Wide-ResNet-26-10&26.75&15.33(2.2)&15.85&15.85(1.0)&28.47&16.16(2.2)&21.71&21.19(0.8)&39.33&18.09(2.7)&14.55&14.23(0.9)&\textbf{14.23}&\textbf{14.23(1.0)}\\
			& DenseNet-121&42.93&21.77(2.4)&19.11&19.11(1.0)&52.14&24.88(2.5)&28.7&28.95(0.9)&54.5&23.41(2.8)&18.39&18.39(1.0)&\textbf{17.48}&\textbf{17.48(1.0)}\\
			\midrule
			Tiny-ImageNet & ResNet-50&232.85&220.98(1.4)&240.32&238.98(0.9)&234.29&226.29(1.3)&235.04&214.95(0.7)&242.1&240.9(1.3)&204.97&204.97(1.0)&\textbf{203.3}&\textbf{203.3(1.0)}\\
			\bottomrule
		\end{tabular}%
	}
	
	\caption{\textbf{NLL $(\%)$ evaluated for different methods.}  Both pre and post temperature scaling results are reported. Optimal temperature is included in brackets (calculated temperature on best ECE). }
	\label{table:nll_tab1}

\end{table*}

\begin{table*}[!t]
	\centering
	\scriptsize
	\resizebox{\linewidth}{!}{%
		\begin{tabular}{cccccccccccccccc}
			\toprule
			\textbf{Dataset} & \textbf{Model} & \multicolumn{2}{c}{\textbf{Weight Decay}} &
			\multicolumn{2}{c}{\textbf{Brier Loss}} & \multicolumn{2}{c}{\textbf{MMCE}} &
			\multicolumn{2}{c}{\textbf{Label Smoothing}} & \multicolumn{2}{c}{\textbf{Inverse Focal Loss}} &
			\multicolumn{2}{c}{\textbf{Focal Loss}} &
			\multicolumn{2}{c}{\textbf{Dual Focal}} \\
			\textbf{} & \textbf{} &
			\multicolumn{2}{c}{\citep{guo2017calibration}} &
			\multicolumn{2}{c}{\citep{brier1950verification}} & \multicolumn{2}{c}{\citep{kumar2018trainable}} &
			\multicolumn{2}{c}{\citep{szegedy2016rethinking}} & \multicolumn{2}{c}{\citep{wang2021rethinking}} &
			\multicolumn{2}{c}{\citep{mukhoti2020calibrating}} &
			\multicolumn{2}{c}{Ours} \\
			&& Pre T & Post T & Pre T & Post T & Pre T & Post T & Pre T & Post T & Pre T & Post T & Pre T & Post T & Pre T & Post T \\
			\midrule
			\multirow{2}{*}{CIFAR-10/SVHN} & ResNet-50&94.32&94.56&93.59&93.72&85.17&64.75&78.88&78.89&92.41&92.60&92.48&92.79&94.34&94.34\\
			& DenseNet-121&84.43&81.57&94.65&94.66&85.88&84.87&78.79&78.94&74.08&67.78&89.59&89.59&93.78&93.78\\
			\midrule
			\multirow{2}{*}{CIFAR-10/CIFAR-10-C} & ResNet-50&86.23&86.03&90.21&90.13&89.97&90.11&72.01&72.02&77.81&74.74&89.45&89.56&87.93&87.93\\								
			& DenseNet-121&87.61&86.41&87.38&87.38&84.9&84.88&73.67&73.8&76.72&72.51&89.47&89.47&89.56&89.56\\
			\bottomrule
		\end{tabular}%
	}
	\caption{\textbf{Robustness on Dataset Shift}.\quad AUROC $(\%)$, being the higher the better, is evaluated for different methods with models shifting from CIFAR-10 (in-distribution) to SVHN and CIFAR-10-C as the OoD datasets. }
	\label{table:auroc_tab1}
\end{table*}

\section{Comparison with SOTA Method KDE-XE}
KDE-XE~\cite{popordanoska2022consistent} is a tractable, differentiable, and consistent estimator of the expected Lp canonical calibration error based on the Dirichlet kernel. For fair comparison, we use the same training setting provided in~\cite{popordanoska2022consistent}  and evaluate the calibration performance on ECE and L1 canonical calibration error. We report the calibration performance on four different models (ResNet-110, ResNet-110sd, Wide-ResNet and DenseNet-40) and two datasets (CIFAR10 and CIFAR100) in Table~\ref{table:KDE-XE}. The results indicate that our methods outperform KDE-XE in terms of ECE. However, Dual focal loss fails to outperform KDE-XE in terms of canonical calibration error. We argue that KDE-XE takes the L1 canonical calibration error as an auxiliary training loss, which makes it difficult to outperform in terms of canonical calibration error. We contend that while canonical calibration error is a significant metric, traditional evaluation metrics such as ECE are still commonly accepted in recent studies, such as ~\cite{mukhoti2020calibrating, ghosh2022adafocal}. Our approach demonstrates a substantially superior ECE performance compared to KDE-XE.

\begin{table}[ht]
\centering
\scriptsize
\resizebox{0.6\linewidth}{!}{%
\begin{tabular}{cccccc}
\toprule
\textbf{Metrics} &\textbf{Model} &\textbf{Dataset} & \textbf{Cross Entropy} & \textbf{KDE-XE} & \textbf{DualFocal}\\

\midrule
ECE & ResNet-110 & CIFAR-10  & 3.89	&3.093	&0.4\\
ECE & ResNet-110sd & CIFAR-10  & 3.555	&2.778	&1.51\\
ECE & ResNet-110 & CIFAR-100  & 12.769	&8.969	&2.43\\
ECE & ResNet-110sd & CIFAR-100  & 11.175	&7.828	&2.6\\
ECE & Wide-ResNet & CIFAR-100  & 7.279	&3.703	&1.04\\
ECE & DenseNet-40 & CIFAR-100  & 9.196	&8.016	&2.34\\
$\text{ECE}_{\text{KDE}}$ & ResNet-110 & CIFAR-10  & 0.133	&0.126	&0.153\\
$\text{ECE}_{\text{KDE}}$ & DenseNet-40 & CIFAR-10  & 0.104	&0.098	&0.121\\
\bottomrule
\end{tabular}}
\caption{\textbf{Comparison with SOTA Method KDE-XE.}}
\label{table:KDE-XE}
\end{table}

\section{Calibration Assessment with Metrics RBS}
KDE-RBS~\cite{gruber2022better} is a calibration metric proposed recently, which is robust w.r.t. the test set size. To evaluate the performance of our method on RBS, we conduct experiments on two datasets (cifar10 and cifar100) and four models (Resnet50, Resnet110, Densenet121 and Wide-Resnet), and our results demonstrate a consistent enhancement over other methods in terms of RBS. Table~\ref{table:RBS} shows the RBS performance of different methods.

\begin{table*}[ht]
	\centering
	\scriptsize
	\resizebox{\linewidth}{!}{%
		\begin{tabular}{ccccccccc}
			\toprule
			\textbf{Dataset} & \textbf{Model} & \textbf{Weight Decay} &
			\textbf{MMCE} & \textbf{Brier Loss} & 
			\textbf{Label Smoothing} &
			\textbf{Focal Loss} &
			\textbf{Dual Focal} \\
			\textbf{} & \textbf{} &
			\citep{guo2017calibration} &
            \citep{kumar2018trainable} &
			\citep{brier1950verification} & 
			\citep{szegedy2016rethinking} & 
			\citep{mukhoti2020calibrating} &
			Ours \\ 
			\midrule
			
			\multirow{4}{*}{CIFAR-10} & ResNet-50
			&0.0922&0.0938&0.0815&0.0943&0.0801&0.0774\\
			&ResNet-110&0.0916&0.1034&0.0879&0.1013&0.0839&0.0798\\
			& Wide-ResNet-26-10&0.0691&0.071&0.0655&0.0728&0.0634&0.0624\\
			& DenseNet-121&0.0929&0.1034&0.0817&0.0949&0.0842&0.081\\
			\midrule
			\multirow{4}{*}{CIFAR-100} &ResNet-50&0.3974&0.3748&0.3424&0.3574&0.3315&0.3193\\
			& ResNet-110&0.4071&0.4123&0.3725&0.3774&0.3366&0.322\\
			& Wide-ResNet-26-10&0.3519&0.3391&0.2966&0.3171&0.2852&0.2848\\
			& DenseNet-121&0.4457&0.4181&0.339&0.4003&0.3218&0.3157\\
			\bottomrule
		\end{tabular}%
	}
	
	\caption{\textbf{Calibration Assessment with Metrics RBS}}
	\label{table:RBS}

\end{table*}

\section{Robustness of DFL}
To verify the robustness of our method, we replicated the experiment using Resnet50 on CIFAR10 and presented the mean calibration error along with the standard error in Table~\ref{table:robust}. These results indicate that our approach exhibits a consistent and robust calibration performance.
\begin{table*}[ht]
	\centering
	\scriptsize
	\resizebox{\linewidth}{!}{%
		\begin{tabular}{lccccccc}
			\toprule
			\textbf{} & \textbf{Weight Decay} &
			\textbf{MMCE} & \textbf{Brier Loss} & 
			\textbf{Label Smoothing} &
			\textbf{Inverse Focal Loss} &
			\textbf{Focal Loss} &
			\textbf{Dual Focal} \\
			\textbf{} &
			\citep{guo2017calibration} &
            \citep{kumar2018trainable} &
			\citep{brier1950verification} & 
			\citep{szegedy2016rethinking} & 
            \citep{wang2021rethinking} & 
			\citep{mukhoti2020calibrating} &
			Ours \\ 
			\midrule
			
			ECE
			&4.4±0.15&1.81±0.07&4.51±0.06&3.01±0.13&4.45±0.15&1.45±0.05&0.49±0.05\\
			ECE(T)&1.37±0.16&1.21±0.15&1.18±0.05&1.66±0.06&1.33±0.07&0.99±0.08&0.49±0.05\\
			AdaECE&4.37±0.17&1.8±0.1&4.51±0.07&2.99±0.09&4.39±0.11&1.49±0.1&0.47±0.04\\
			Classwise-ECE&0.93±0.3&0.41±0.05&0.94±0.03&0.72±0.05&0.95±0.04&0.42±0.01&0.35±0.01\\
			\bottomrule
		\end{tabular}%
	}
	
	\caption{\textbf{Robustness of DFL}}
	\label{table:robust}

\end{table*}

\section{Gamma Value Selection in DFL}
We simply discover the appropriate $\gamma$ via cross-validation, which is a normal technique as mentioned in ~\cite{mukhoti2020calibrating}, “Finding an appropriate $\gamma$ is normally done using cross-validation. Also, traditionally, $\gamma$ is fixed for all samples in the dataset.” We notice that the FLSD-53 strategy is utilized in ~\cite{mukhoti2020calibrating} to better control the gradient magnitude via a better trade-off in the curve of function g which is discussed in ~\cite{mukhoti2020calibrating}. The major reason why we use fixed $\gamma$ in DFL instead of FLSD-53 strategy lies in the fact that DFL naturally performs gradient magnitude control and better trade-offs in terms of function g. 

We agree that adjusting $\gamma$ can be an effective solution to the better calibration performance with focal loss. However, our proposed DFL can also meet the requirements in ~\cite{mukhoti2020calibrating} through the involvement of other logits. We further provide more empirical results of $\gamma$ selection on ResNet50 on CIFAR10 in Table~\ref{table:gamma}.

\begin{table*}[ht]
	\centering
	\scriptsize
	\resizebox{0.5\linewidth}{!}{%
		\begin{tabular}{lccccc}
			\toprule
   				
			\textbf{} & \textbf{Accuracy} &
			\textbf{ECE} & \textbf{ECE(T)} & 
			\textbf{AdaECE} &
			\textbf{Classwise ECE}\\
			\midrule
			
			$\gamma = 2$
			&95.3&2.156&0.8553&2.144&0.5167\\
			$\gamma = 3$&94.69&1.79&1.179&1.991&0.4828\\
			$\gamma = 4$&94.75&1.617&1.024&1.69&0.4958\\
			$\gamma = 5$&94.83&0.461&0.461&0.6694&0.3578\\
			$\gamma = 10$&94.44&1.096&1.096&1.446&0.4406\\
			\bottomrule
		\end{tabular}%
	}
	
	\caption{\textbf{Gamma Selection}}
	\label{table:gamma}

\end{table*}

\end{document}